\documentclass[11pt,final]{amsart}

\usepackage{comment}
\usepackage{booktabs}       
\usepackage{amsfonts}       
\usepackage{nicefrac}       
\usepackage[noend]{algpseudocode}
\usepackage{amsmath,amsfonts,amssymb,amscd}
\usepackage[foot]{amsaddr}
\usepackage{lastpage}
\usepackage{enumerate}
\usepackage{fancyhdr}
\usepackage{mathrsfs}
\usepackage[table]{xcolor}
\usepackage{graphicx}
\usepackage{listings}
\usepackage{hyperref,leftidx}
\usepackage{bbm}
\usepackage{subfig}
\usepackage{caption}
\usepackage{diagbox,multirow}
\usepackage[toc,page]{appendix}

\usepackage{tikz-qtree,tikz-qtree-compat}
\usepackage{tikz,pgf}
\usepackage{arydshln}
\usepackage{booktabs}
\usepackage[margin=0.75in]{geometry}

\usetikzlibrary{decorations,decorations.markings,decorations.text}
\usetikzlibrary{arrows.meta}

\usepackage{listings}
\usepackage{hyperref}
\usepackage{bbm}
\usepackage{subfig}
\usepackage{caption}
\usepackage{diagbox} 

\usepackage{arydshln}
\usepackage{booktabs}

\usepackage[displaymath, mathlines]{lineno}

\usetikzlibrary{decorations,decorations.markings,decorations.text}
\usetikzlibrary{arrows.meta}
\usetikzlibrary{patterns}

\usetikzlibrary{positioning,patterns}
\usetikzlibrary{calc,fadings,decorations.pathreplacing,arrows,
datavisualization.formats.functions,shapes.geometric}

\hypersetup{
    colorlinks=true,
    linkcolor=blue,
    filecolor=blue,      
    urlcolor=blue,
    citecolor = blue,
}

\def\boxit#1{\vbox{\hrule\hbox{\vrule\kern6pt
          \vbox{\kern6pt#1\kern6pt}\kern6pt\vrule}\hrule}}

\def\bse{\begin{eqnarray*}}
\def\ese{\end{eqnarray*}}
\def\be{\begin{eqnarray}}
\def\ee{\end{eqnarray}}
\def\bq{\begin{equation}}
\def\eq{\end{equation}}
\def\bse{\begin{eqnarray*}}
\def\ese{\end{eqnarray*}}

\def\T{^{\rm T}}

\definecolor{darkblue}{rgb}{0,0.08,0.4}
\definecolor{brightblue}{rgb}{0.65,0.85,0.85}
\definecolor{darkred}{rgb}{0.8,0.2, 0.2}
\definecolor{darkgreen}{rgb}{0, 0.6, 0}
\definecolor{blueish}{rgb}{0.1176, 0.5647, 1.0000}
\definecolor{darkorange}{RGB}{255,140,0}
\definecolor{babyblue}{RGB}{153,204,255}

\definecolor{colorone}{rgb}{0.1176,0.5647,1.0000}
\definecolor{colortwo}{rgb}{0.5608,0.7373,0.5608}

\definecolor{mlsbg}{HTML}{EAF1F7}

\newcommand{\corb}[1]{\textcolor{black}{#1}}

\newcommand{\rev}[1]{\textcolor{black}{#1}}
\newcommand{\revt}[1]{\textcolor{black}{#1}}

\newcommand{\jc}[1]{\textcolor{black}{#1}}

\newcommand{\ks}[1]{\textcolor{black}{#1}}

\newcommand{\bbN}{\mathbb{N}}
\newcommand{\bbE}{\mathbb{E}}

\newcommand{\bbP}{\mathbb{P}}

\newcommand{\bC}{\mathbf{C}}
\newcommand{\bA}{\mathbf{A}}
\newcommand{\bB}{\mathbf{B}}

\newcommand{\bK}{\mathbf{K}}

\newcommand{\bx}{\mathbf{x}}
\newcommand{\by}{\mathbf{y}}

\newcommand{\bw}{\mathbf{w}}
\newcommand{\bm}{\mathbf{m}}

\newcommand{\mcA}{{\mathcal A}}
\newcommand{\mcB}{{\mathcal B}}

\newcommand{\mcF}{\mathcal{F}}

\newcommand{\mcM}{{\mathcal M}}
\newcommand{\mcN}{{\mathcal N}}

\newcommand{\eset}[1]{{\mathbb E} \left[ #1 \right] }

\newcommand{\Imag}{\mathop{\text{\rm Im}}}

\newcommand{\Span}[1]{\mathrm{Span}\{ #1 \}}

\def\R{\Bbb{R}}

\newtheorem{defi}{Definition}
\newtheorem{theorem}{Theorem}

\newtheorem{lemma}{Lemma}

\theoremstyle{remark}
\newtheorem{asum}{Assumption}

\newtheorem{remark}{Remark}

\DeclareMathOperator{\spn}{span}

    \usetikzlibrary{arrows.meta,
                    bending,
                    calc,chains,
                    shapes.misc,
                    babel
                    }

\tikzstyle{blockTwo} = [rounded corners=1ex,fill=green!40!blue!20!white!50!,minimum width=2em]
\tikzstyle{emptyblock} = [draw,minimum width=2em]
 
\tikzstyle{branch}=[fill,shape=circle,minimum size=3pt,inner sep=0pt]
 \tikzstyle{vecArrowTwo} = [fill=blue!40,blue!40,thick,
decoration={markings,mark=at position 1 with
   {\arrow[blue!40]{triangle 60}}}, double distance=1.4pt,
   shorten >= 5.5pt, preaction = {decorate}, postaction = {draw, 
   blue!40,shorten >= 4.5pt}]

\pgfmathdeclarefunction{elvsin}{2}{%
  \pgfmathparse{(1/(#2*sqrt(2*pi))*exp(-((x-#1)^2)/(2*#2^2)))*(sin(50*pi*x)
    + 1)}
}%

\title{Stochastic tensor space feature theory with applications to robust machine learning}

\author{Julio E. Castrill\'{o}n-Cand\'{a}s$^{\ddagger}$, Kaili
  Shi$^{\ddagger}$, Trajan Murphy $^{\ddagger}$, Dingning Liu$^{\ddagger}$, Sicheng
  Yang$^{\ddagger}$, Xiaoling Zhang$^{\dagger}$, Mark
  Kon$^{\ddagger}$, the Alzheimer's
  Disease Neuroimaging Initiative$^{*}$}

\email{%
  \parbox{0.85\linewidth}{\raggedright
    jcandas@bu.edu, shikl@bu.edu, trajanm@bu.edu, dnzliu@ucdavis.edu,
    sichengy@bu.edu,\\[-1pt]
    zhangxl@bu.edu, mkon@bu.edu
  }%
}

 \address{
   ${\ddagger}$ Department of Mathematics and Statistics, 
  Boston University, Boston, MA.  ${\dagger}$ School of Medicine , Boston 
  University, Boston, MA.
  }

\thanks{$^{*}$Data used in preparation of this article were obtained
  from the Alzheimer’s Disease Neuroimaging Initiative (ADNI) database
  (adni.loni.usc.edu). As such, the investigators within the ADNI
  contributed to the design and implementation of ADNI and/or provided
  data but did not participate in analysis or writing of this
  report. A complete listing of ADNI investigators can be found at:
  \href{http://adni.loni.usc.edu/wp-content/uploads/how_to_apply/ADNI_Acknowledgement_List.pdf}{http://adni.loni.usc.edu/wp-content/uploads/how\_to\_apply/ADNI\_Acknowledgement\_List.pdf}}


\begin{document}

\maketitle

\begin{abstract}
  In this paper we develop a Multilevel Orthogonal Subspace (MOS) Karhunen--Lo\`{e}ve feature theory based on stochastic tensor spaces, for the construction of robust machine learning features. Training data are treated as instances of a random field within a relevant Bochner space. Our key observation is that separate machine learning classes can reside predominantly in mostly distinct subspaces. Using the Karhunen-Lo\`{e}ve expansion and a hierarchical expansion of the first (nominal) class, a MOS is constructed to detect anomalous signal components, treating the second class as an outlier of the first. The projection coefficients of the input data into these subspaces are then used to train a Machine Learning (ML) classifier. These coefficients become new features from which much clearer separation surfaces can arise for the underlying classes. We test our approach on blood plasma and cerebrospinal fluid datasets (Alzheimer’s Disease Neuroimaging Initiative).  The tests show significant increases in accuracy by applying them to Support Vector Machines and (Deep) Neural Networks. Furthermore, from non-invasive blood test, high-accuracy results can be obtained for predicting AD stages such as cognitive normal, mild cognitive impairment and dementia. The non-invasive nature of this test is significant since it can potentially avoid other painful tests.
\end{abstract}



In the past decades there has been an emphasis on the development of more accurate Machine Learning (ML) algorithms.  Algorithms such as Deep Neural Networks (DNNs) have become increasingly complex and difficult to interpret mathematically, and therefore hard to construct and optimize. Despite the success of DNNs, the development process is largely labor-intensive and based on trial and error, in particular for DNNs with their large numbers of parameters and data-intensive training needs. We take a fundamentally different approach to ML and develop a stochastic dynamic tensor theory for the construction of robust machine learning features that can significantly speed up the process of building good machine learning algorithms. Furthermore, our approach also significantly improves the performance of standard machine learning algorithms, including Neural Networks.

We introduce a systematic approach for the construction of ML feature vectors that improves class separations, using techniques in probability theory and the recent Functional Data Analysis (FDA) theory on anomaly detection.  Our implementations involve techniques from computational applied mathematics and computer science. We apply this to the well-known Alzheimer’s Disease Neuroimaging Initiative (ADNI) blood plasma proteomics dataset \cite{Petersen2010} for cognitive impairment classification, with very significant increases in accuracy. Due to the non-invasive nature of blood tests, our results are highly significant from a medical perspective. In contrast, the cerebrospinal fluid (CSF) proteomic test, which is considered the best method for live AD detection, is highly invasive and painful. In addition we test our method on CSF and AD gene expression data (newAD).


Due to its foundation in functional analysis and tensor product expansions, this approach can be easily extended to classification problems on complex topologies, including gene expression networks.

\emph{We treat the data as realizations of a random field in a suitable Bochner space. The key insight of our approach is that classes can be often localized to two separate subspaces in a Bochner function space. By exploiting the Karhunen-Lo\`{e}ve expansion, these subspaces can be constructed to reveal an optimal class separation (See Figure \ref{intro:fig1}}).  This paper is a novel application of the theory developed in \cite{Castrillon2022}. In particular, we show how functional data analysis \cite{Horvath2012,Kokoszka2017} can be applied to statistical ML.  The problem of classification in ML has been studied and benchmarked for decades, and this method provides an entirely new way of approaching the problem with a new feature map that is based on novel estimates of the underlying covariance structure, applied to quantitative anomaly calibrations as novel ML features. This approach effectively augments current ML algorithms.

The Karhunen--Lo\`{e}ve (KL) expansion is strongly related to Principal Component Analysis (PCA). In the discrete setting they are practically the same. PCA is widely used for building ML features by using the principal components. However, most applications of PCA tend to ignore the underlying probabilistic interpretation. In contrast, by using the KL expansion of random fields (or random vectors for the discrete case) and the theory developed in this paper, we conclude that it is not the principal components but rather the residual eigenspaces which are important for classification. We explore this in detail below.

A fundamental issue in ML predictive modeling is robustness and sensitivity to data quality.  ML with complex noisy observations involves a host of difficulties including the problem of overfitting, which can give rise to highly unstable and inaccurate decision boundaries.  This problem is particularly difficult for data with high dimension ($p$) and low sample size ($N$), (i.e., $p \gg N$), for example genome-wide gene expression data, in which the number of genes is in the tens of thousands while available samples are limited by the high cost of high-throughput profiling assays and limited access to tissue samples.  In addition, the problem can also present itself in high dimension with even larger sample sizes (i.e. $p \ll N$), for example, as in the UK-Biobank dataset ($N = ~500,000$), with noisy inputs again leading to unstable decision boundary oscillations. Such oscillations generally result from overfitting noise, and can lead to poor machine performance.

In Figure \ref{intro:fig1} (a), (b), (c) an illustrative example of
  such classification is shown. In (a), the data are well separated,
  with blue dots representing the first class and orange dots the
  second. Due to the separation of the data it is in principle easy to
  construct a decision boundary. In (b) the data classes are
  mixed. Furthermore, the data can be noisy and diffusive in high
  dimensions, leading to unstable boundary decision surfaces. The
  fundamental questions are: does there exist a good separation of the
  data classes in some coordinate system? How can we construct a
  transformation revealing this separation in an appropriate space
  (See Figure \ref{intro:fig1}(c))?

\begin{figure}
  \begin{center}
    \begin{tikzpicture}[scale = 0.75,every node/.style={scale=0.77}]>=latex']
    \begin{scope}[xshift = -15cm, font=\normalsize,
     dotout/.style={circle,draw=blueish,fill=blueish, inner
       sep=0pt,minimum size=1.5mm},
     dotin/.style={circle,draw=darkorange,fill=darkorange, inner
       sep=0pt,minimum size=1.5mm}]
   \draw [-stealth] (-2.75, 0) -- (2.75, 0) node [below left]  {\normalsize };
   \draw [-stealth] (0,-2.75) -- (0,2.75) node [below left] {\normalsize };
   \draw [dashed] (-2,2) -- (2,-2) node [below left] {\normalsize };   
    \node at (0,   -3.5) [] {(a) Classification (Easy)};
    \node at (0.0066,   0.2705) [dotin] {};
    \node at (1.2249,   1.2646     ) [dotin] {};
    \node at (1.6408,    0.2353) [dotin] {};
    \node at (0.3640,    1.0547) [dotin] {};
    \node at (1.7070,    1.7853) [dotin] {};
    \node at (0.5740,   1.8433) [dotin] {};
    \node at (1.2360,    0.9966) [dotin] {};
\node at (-0.7586,-0.1541) [dotout] {}; 
\node at (-1.4727,-0.8712) [dotout] {}; 
\node at (-1.5060,-1.9038) [dotout] {}; 
\node at (-1.2364,-1.8543) [dotout] {}; 
\node at (-1.4171,-0.6950) [dotout] {}; 
\node at (-0.3402,-1.8921) [dotout] {}; 
\node at (-1.7066,-0.5399) [dotout] {}; 
\node at (-1.2893,-0.7634) [dotout] {}; 
\node at (-0.8916,-1.5164) [dotout] {}; 
\node at (-1.7034,-1.6631) [dotout] {}; 
\node at (-0.0294,-1.4581) [dotout] {}; 
\node at (-0.6195,-0.6715) [dotout] {}; 
\node at (-1.5605,-1.5899) [dotout] {}; 
       \end{scope}
    \begin{scope}[xshift = -8cm, font=\normalsize,,
     dotout/.style={circle,draw=blueish,fill=blueish, inner
       sep=0pt,minimum size=1.5mm},
     dotin/.style={circle,draw=darkorange,fill=darkorange, inner
       sep=0pt,minimum size=1.5mm}]
   \draw [-stealth] (-2.75, 0) -- (2.75, 0) node [below left]  {\normalsize };
    \draw [-stealth] (0,-2.75) -- (0,2.75) node [below left] {\normalsize };
    \node at (0,   -3.5) [] {(b) Classification (Hard)};   
    \node at (-0.0066,   -0.2705) [dotin] {};
    \node at (-1.2249,   -1.2646     ) [dotin] {};
    \node at (1.6408,    0.2353) [dotin] {};
    \node at (0.3640,    1.0547) [dotin] {};
    \node at (-1.7070,    1.7853) [dotin] {};
    \node at (-0.5740,   -1.8433) [dotin] {};
    \node at (1.2360,    0.9966) [dotin] {};

\node at (0.7586,0.1541) [dotout] {}; 
\node at (-1.4727,-0.8712) [dotout] {}; 
\node at (-1.5060,1.9038) [dotout] {}; 
\node at (-1.2364,-1.8543) [dotout] {}; 
\node at (-1.4171,-0.6950) [dotout] {}; 
\node at (0.3402,1.8921) [dotout] {}; 
\node at (-1.7066,-0.5399) [dotout] {}; 
\node at (1.2893,-0.7634) [dotout] {}; 
\node at (0.8916,-1.5164) [dotout] {}; 
\node at (1.7034,1.6631) [dotout] {}; 
\node at (-0.0294,-1.4581) [dotout] {}; 
\node at (0.6195,-0.6715) [dotout] {}; 
\node at (1.5605,1.5899) [dotout] {}; 

\coordinate (O) at (3.5,0);
\coordinate (P) at (4.8,0);
\draw[->, >=latex, gray, line width=4 pt] (O) -- (P);
    \end{scope}
    \begin{scope}[font=\normalsize,
     dotout/.style={circle,draw=blueish,fill=blueish, inner
       sep=0pt,minimum size=1.5mm},
     dotin/.style={circle,draw=darkorange,fill=darkorange, inner
       sep=0pt,minimum size=1.5mm}]        
    \filldraw[fill=green!40!blue!20!white!50!,dashed] (0,0) ellipse (2 and 2);       
    \node at (-0.5,-0.5) [dotin] {};
    \node at (0.2,-0.3) [dotin] {};
    \node at (0.5,0.5) [dotin] {};
    \node at (-0.25,0.25) [dotin] {};
    \node at (0.25,0.25) [dotin] {};
    \node at (-0.35,-0.25) [dotin] {};
    \node at (1.5,2) [dotin] {};
    \node at (2,2) [dotout] {};
    \node at (2,1.5) [dotout] {};
    \node at (2.25,2) [dotout] {};
    \node at (2.25,2.25) [dotout] {};

    \node at (-2,1) [dotout] {};
    \node at (-2.25,1) [dotout] {};
    \node at (-2,1.25) [dotout] {};
    \node at (-2.35,2.35) [dotout] {};

    \node at (1,-2) [dotout] {};
    \node at (1.25,-2) [dotout] {};
    \node at (1,-2.25) [dotout] {};
    \node at (0.75,-2) [dotout] {};

    \node at (-0.25,0.5) [dotout] {};

    \node at (-0.25,-0.5) [dotout] {};
    \draw [-stealth] (-2.75, 0) -- (2.75, 0) node [below left]  {};
    \draw [-stealth] (0,-2.75) -- (0,2.75) node [below left] {};
    \node at (0,   -3.5) [] {(c) Mapped feature};
    \end{scope}
\end{tikzpicture}
  \end{center} \caption{Illustration of binary classification with
  classes denoted by blue and orange points.  In (a) we see that the
  data are well separated, with blue dots forming the first class and
  orange dots the second class. Due to the separation of the data it
  is in principle easy to construct a decision boundary. (b) For this
  case the data classes are mixed, leading to complex boundary
  decision surfaces that are hard to build, yielding low
  accuracy. Furthermore the data can be noisy and diffusive in high
  dimensions, leading to unstable boundary decision surfaces. (c)
  After applying an appropriate transformation using stochastic
  coordinate transformations the classes separate, leading to stable
  boundary decision surfaces.}  \label{intro:fig1}
\end{figure}

By applying an appropriate transformation, the separation between the
classes is revealed. For example, consider the temporal functions \(
f_{\bA}(t) = \sin{t} \) and \( f_{\bB}(t) = \sin{2t} \). Given an
observation \( f(t) = f_{\bA}(t) + f_{\bB}(t) \), we aim to separate
the components into the two classes \( f_{\bA}(t) \) and \(
f_{\bB}(t) \).  In the time domain, performing the separation is
harder directly using unsupervised approaches, but in an appropriate
mapped image space this can be easily done.  Performing the Fourier
transform on $f_{\bA}(t)$ and $f_{\bB}(t)$, we obtain $f_{\bA}(\xi) =
i(\delta(\xi+1) - \delta(\xi - 1))$ and $f_{\bB}(\xi) =
i(\delta(\xi+2) - \delta(\xi - 2))$. Thus from the Fourier transform
of $f(t)$ we can easily distinguish the signals from \(
f_{\bA}(\xi) \) and \( f_{\bB}(\xi) \), making classification much
easier. (See Figure \ref{fourier}).

\tikzset{>=latex} 
\colorlet{myblue}{blue!80!black}
\colorlet{mydarkblue}{myblue!80!black}
\tikzstyle{xline}=[myblue,thick]
\def\tick#1#2{\draw[thick] (#1) ++ (#2:0.1) --++ (#2-180:0.2)}
\tikzstyle{myarr}=[myblue!50,-{Latex[length=3,width=2]}]
\def\N{80}
\begin{figure}[t]
\centering
\def\xmin{-0.7*\T} 
\def\xmax{3.0}     
\def\ymin{-0.4}    
\def\ymax{1.7}     
\def\A{0.67*\ymax} 
\def\T{0.31*\xmax} 
\begin{tikzpicture}
  \def\T{0.30*\xmax} 
  \def\A{0.70*\ymax} 
  \draw[->,thick] (0,\ymin) -- (0,2) node[above=0] {$f(t)$};
  \draw[->,thick] (-\xmax,0) -- (\xmax+0.1,0) node[below=0] {$t$ };
  \draw[xline,samples=\N,smooth,variable=\t,domain=-0.94*\xmax:0.94*\xmax]
    plot(\t,{\A*  ((sin(360/(\T)*\t) +  (sin(180/(\T)*\t))  )});
  \tick{{-3*\T},0}{90} node[left=  5,below=-1,scale=0.85] {$-3\pi$};
  \tick{{  -\T},0}{90} node[left=  4,below= -1,scale=0.85] {$-\pi$};
  \tick{{   \T},0}{90} node[right= 0,below= -1,scale=0.85] {$\pi$};
  \tick{{ 3*\T},0}{90} node[right=-1,below=-1,scale=0.85] {$3\pi$};
\end{tikzpicture}
\raisebox{1.70mm}{
\begin{tikzpicture}
  \message{^^JRectangular function - frequency domain}
  \def\T{1} 
  \def\A{1} 
  \draw[->,thick] (0,\ymin) -- (0,2) node[above = 0] {$\Imag f(\xi)$};
  \draw[->,thick] (-\xmax,0) -- (\xmax+0.1,0) node[below=0] {$\xi$};
  \tick{{-2*\T},0}{90} node[left=  5,below= -0,scale=0.85] {$-2$};
  \tick{{  -\T},0}{90} node[left=  4,below= -0,scale=0.85] {$-1$};
  \tick{{   \T},0}{90} node[right= 0,below= -0.75,scale=0.85] {$1$};
  \tick{{ 2*\T},0}{90} node[right=-1,below= -0.75,scale=0.85] {$2$};
  \draw[xline,->] (2,0) --++ (0,-\A) node[mydarkblue,below right=1] {};
     \draw[xline,->] (1,0) --++ (0,-\A) node[mydarkblue,below right=1] {};
     \draw[xline,->] (-1,0) --++ (0,\A) node[mydarkblue,below right=1] {};
     \draw[xline,->] (-2,0) --++ (0,\A) node[mydarkblue,below right=1] {};
      \draw[xline,thick,fill=white] (-2,0) circle(0.05);
      \draw[xline,thick,fill=white] (-1,0) circle(0.05);
      \draw[xline,thick,fill=white] (1,0) circle(0.05);
      \draw[xline,thick,fill=white] (2,0) circle(0.05);
\end{tikzpicture}}
\caption{Coordinate transformation reveals the frequency components of the signal 
$f(t)$ thus making it easier to classify and distinguish.  These plots
are created in TikZ by modifying the LaTex code
from \cite{Neutelings2021a,Neutelings2021b}.}
\label{fourier}
\end{figure}
    

\revt{Our approach is to treat the data as if they are realizations of a random field in an appropriate Bochner tensor product space. By using a suitable stochastic coordinate system the separation between the signals can be revealed.}
Our overarching goals in this work are: i) to develop stochastic
functional (data) analysis approaches for significantly
improving accuracy and robustness of ML methods on high
dimensional noisy datasets (which can also be based on complex
topologies); ii) to motivate and develop high performance computing
algorithms based on these approaches.

The related signal decomposition is an exact hierarchical tensor
product expansion with known optimality properties for approximating
stochastic processes (random fields) with finite dimensional function
spaces as ranges.  In principle, these primary low dimensional range
spaces can capture most of the stochastic behavior of underlying
signals in a given nominal class, and can reject signals in
alternative classes as stochastic anomalies.  Using a hierarchical
finite dimensional KL expansion for the nominal class, a series of
orthogonal nested subspaces is constructed for detecting anomalous
signal components relative to the nominal class.  Projection
coefficients of input data in these subspaces are then used to train
an ML classifier. However, due to the split of the signal into nominal
and anomalous projection components, clearer separation surfaces of
the classes arise. In fact, we show that with a sufficiently accurate
estimation of the covariance structure of the nominal class, a sharp
classification can be obtained. This is particularly advantageous when
large imbalanced datasets are available.

\ks{We formulated this framework and demonstrated its effectiveness across several high-dimensional datasets. The MOS-KL representation can substantially improve classification performance relative to standard machine-learning methods trained on the original features. The largest improvement is observed for the ADNI blood-plasma proteomics task distinguishing cognitively normal (CN) participants from those with mild and late cognitive impairment (MCI/LMCI). Despite the pronounced class imbalance (54 CN and 346 MCI/LMCI samples), an RBF-kernel Support Vector Machine (SVM) \cite{Scholkopf1997} trained on the MOS-KL features achieves 96.3\% accuracy, compared with 91.0\% for the same SVM trained on the original features and 91.3\% for the strongest ensemble baseline among Gradient Boosting \cite{Friedman2002}, RUSBoost \cite{Seiffert2008}, and Random Forest \cite{Ho1995}. The corresponding MOS-KL SVM also provides a well-balanced sensitivity of 94.8\% and specificity of 97.7\%. By contrast, classifiers trained on the original features exhibit sensitivity--specificity disparities of approximately 15--66 percentage points. The benefits of MOS-KL extend beyond predictive performance: across the ADNI plasma, CSF, and newAD cohorts, the mean fit-and-predict time is reduced by approximately 25\% for the SVM and 32\% for the multilayer perceptron (MLP). These results indicate that MOS-KL can simultaneously improve classification accuracy, mitigate majority-class bias, and reduce computational cost.} 


\begin{remark}
The SVM model is created using the fitcsvm command in
MATLAB \cite{Matlab2025}. In this paper we apply the fitposterior
command to the SVM model. This transformation makes the SVM algorithm
more robust to imbalanced datasets and more accurate.  The
fitposterior score-to-posterior-probability transformation function
algorithm is implemented from \cite{Platt2000,Tao2005}.
\end{remark}


Highly imbalanced datasets can be a difficult problem for ML
algorithms. There are many approaches to compensate for an imbalanced
dataset. This may involve, for example, removing the imbalanced
portion of the data, bootstrapping to create more samples of the
smaller class of data, or adjusting the ML algorithm by using
weights \cite{Xanthopoulos2014}. However, many of these solutions are
unsatisfactory.  In particular, if the number of samples of the
smaller class is very small and/or are noisy.  This has motivated the
development of one class semi-supervised methods. See
\cite{perera2021} for a comprehensive survey of these methods.
In this study, we develop the Multilevel Orthogonal Subspace (MOS) KL
feature theory (or Multilevel features for short) to solve this
problem in a more elegant form. Furthermore, as the dataset becomes
more imbalanced the accuracy of our approach increases. We do note
that modern methods such as RUS Boost are also robust to imbalanced
datasets; however, the MOS-KL approach still significantly surpasses
RUS Boost for the series of tests that we have performed.


In the appendix we demonstrate the power of the MOS-KL method by applying it
to a series of tests are performed on imbalanced semi-synthetic datasets created from the GCM
dataset \cite{Ramaswamy2001}. These tests show that our MOS features
approach is robust and performs well under highly imbalanced
datasets. It not only outperforms popular ML methods such as Neural Network, Random
Forest, SVM, Gradient Boosting, which are susceptible to imbalanced
datasets, but also RUS Boost and SVM (with a posterior fit) methods
that are also robust to imbalanced datasets.  Furthermore, tests on
complex imbalanced semi-synthetic data in the appendix show that the
increase of available data on the large class \emph{dramatically} improves accuracy.

\section{Methods}  
\subsection{Mathematical preliminaries}

We demonstrate our approach to the ML classification problem.  
A novel strategy for classification will be demonstrated here via
construction of a series of subspaces orthogonal to a stochastic
representation of data belonging to one of the classes. For two class
classification, the second class is treated as a change or anomaly
with respect to the first. The constructed subspaces allow detection
of such `anomalies' with high accuracy from data and projection
coefficients, and contain the information used to train an ML
classifier.

More precisely, the variations of the data are viewed in terms of a realization
of a random field; the Karhunen Lo\`{e}ve expansion is an important
tool for representing such fields as spatial-stochastic tensor
expansions. This optimal decomposition is well suited for the analysis
of such random fields.  Let $(\Omega,\mcF,\bbP)$ be a complete
probability space, with $\Omega$ a set of outcomes, and $\mcF$ a
$\sigma$-algebra of events equipped with the probability measure
$\bbP$.  Let $U$ be a domain of $\R^{d}$ and $L^{2}(U)$ be the Hilbert
space of all square integrable functions $v:U \rightarrow \R$ equipped
with the standard inner product $\langle u,v \rangle = \int_{U} uv\,
\mbox{d}\bx,$ for all $u(\bx),v(\bx) \in L^{2}(U)$.  In addition, let
$L^{2}_{\bbP}(\Omega; L^{2}(U))$ be the space of all functions
$v:\Omega \rightarrow L^{2}(U)$ equipped with the inner product
$\langle u,v \rangle_{L^{2}_{\bbP}(\Omega; L^{2}(U))} = \int_{\Omega}
\langle u,v \rangle \, \mbox{d}\bbP,$ for all $u,v \in
L^{2}_{\bbP}(\Omega;L^{2}(U))$. \emph{We point out that our approach
is applicable to complex topologies on $U$, including manifolds in
$\R^{d}$, networks, spatio-temporal domains, etc.}

\begin{defi}
Suppose that $v \in L^{2}_{\bbP}(\Omega;L^{2}(U))$.
\begin{enumerate}[i)]
    \item Denote 
    \[
    E_v := \eset{v}: = \int_{\Omega} v(\bx,\omega)\,\mbox{\emph{d}}\bbP
    \]
    as the  mean of $v$.
  \item  Define the covariance function
\[
{\rm Cov}(v(\bx,\omega),v(\by,\omega)) := \bbE[ (v(\bx,\omega) 
- \bbE[v(\bx,\omega)]) (v(\by,\omega) - \bbE[v(\by,\omega)])].
\]
\item Define the linear operator 
$T:L^{2}(U) \rightarrow L^2(U)$ by
\[
T(u)(\bx) := \int_{U}  {\rm Cov}(v(\bx,\omega),v(\by,\omega)) u(\by)\,\mbox{\emph{d}} \by
\]
for all $u \in L^{2}(U)$.  
\end{enumerate}
\label{defn1}
\end{defi}
The above covariance structure will be critical for an accurate
stochastic representation of the random field $v$. In particular, the
eigenstructure of the linear operator $T:L^{2}(U) \rightarrow L^2(U)$
plays a major role. From Lemma 2 and Theorem 1 in
\cite{Harbrecht2016}, there exists a set of eigenfunctions $\{\phi_k\}_{k
  \in \bbN}$, with $\langle \phi_{k}, \phi_{l} \rangle = \delta[i-j]$
and a sequence of eigenvalues $\lambda_1 \geq \lambda_2 \geq \dots>0$
such that $T \phi_k$ = $\lambda_k \phi_k$ for all $k \in \bbN$. From
this eigenstructure the following was proved in Proposition 2.8 in
\cite{Schwab2006}
\begin{theorem}
If $v \in L^{2}(\Omega;L^{2}(U))$, then the random field $v$ can be
represented in terms of the \emph{Karhunen--Lo\`{e}ve} (KL) tensor
product expansion as
\begin{equation}
v(\bx,\omega) = E_v + \sum_{k \in \bbN} \lambda^{\frac{1}{2}}_{k}
\phi_k(\bx) Y_{k}(\omega),
\label{eqn:KL}
\end{equation}
where $\eset{Y_k Y_l} = \delta_{kl}$ and $\eset{Y_k} = 0$ for all $k,l
\in \bbN$.
\label{remote:thm1}
\end{theorem}
From orthogonality properties of the tensor expansion it is not hard
to show that
\[
\| v -  E_v\|^{2}_{L^{2}_{\bbP}(\Omega; L^{2}(U))} = \sum_{k \in \bbN} \lambda_k^{\frac{1}{2}}.
\]
Thus the eigenvalue magnitudes control the contribution to the
variance of each term of the tensor product expansion.

Suppose we are interested in forming the optimal $M$ dimensional
approximation.  We can conclude the optimal choice with respect to the
Bochner norm $\| \cdot \|_{L^{2}_{\bbP}(\Omega; L^{2}(U))}$ is formed
from the first $M$ expansion terms, giving the truncated KL expansion:
\begin{equation}
v_M(\bx,\omega) = E_v + \sum_{k=1}^{M} \lambda^{\frac{1}{2}}_{k}
\phi_k(\bx) Y_{k}(\omega),
\label{KLexpansion}
\end{equation}
with
\[
\| v - v_M\|^{2}_{L^{2}_{\bbP}(\Omega; L^{2}(U))} = \sum_{k = M +
  1}^{\infty} \lambda_k.
\]

In fact it can be shown this is the optimal expansion i.e. no other
orthogonal tensor product expansion has smaller residuals.
From tensor product theory the space $L^{2}_{\bbP}(\Omega;L^{2}(U))$
is isomorphic to $L^{2}_{\bbP}(\Omega) \otimes L^{2}(U)$. In fact it 
can be shown (see \cite{Castrillon2022}) that:

\begin{theorem}
If $\{\pi_i\}_{j=1}^{\infty}$ is a complete orthonormal basis of $L^{2}(U)$ and $\{Z_k\}_{k=1}^{\infty}$ is a complete orthonormal basis of $L^{2}_{\bbP}(\Omega)$ then $\{\{Z_i \pi_j\}_{i=1}^{\infty} \}_{j=1}^{\infty}$ is a complete  orthonormal basis of $L^{2}_{\bbP}(\Omega;L^{2}(U))$.
\end{theorem}

Since $\{\{Z_i \pi_j\}_{i=1}^{\infty} \}_{j=1}^{\infty}$ is a basis
for $L^{2}_{\bbP}(\Omega;L^{2}(U))$ then $v(\bx,\omega)
= \sum_{i,j}^{\infty}
\alpha_{i,j} Z_{i}\phi_{j}$ for some set of coefficients $\alpha_{i,j}$. 
Supposing that we seek an optimal truncated basis to represent the
signal $v(\bx,\omega)$, the KL basis will be optimal. There will be no
$M$-delimited set of tensor product orthonormal functions in
$L^{2}_{\bbP}(\Omega;L^{2}(U))$ that will be better.

Let $H_M
\subset L^{2}(U)$ such that $\dim H_M = M$ and $P_{H_M \otimes
  L^{2}_{\bbP}(\Omega)}: L^{2}(U) \otimes L^{2}_{\bbP}(\Omega)
\rightarrow H_M \otimes L^{2}_{\bbP}(\Omega)$ is an orthogonal
projection operator.  The following theorem is a direct extension of
Theorem 2.7 in \cite{Schwab2006}, showing optimality of KL expansions.

\begin{theorem}
Suppose $f \in L^{2}(U) \otimes L^{2}_{\bbP}(\Omega)$, with $E_f =
0$. Then
\[
\inf_{\begin{array}{c} H_M \subset L^{2}(U) \\ \mbox{dim}\, H_M = M
\end{array}
} \|f - P_{H_M \otimes L^{2}_{\bbP}(\Omega)}f \|_{L^{2}_{\bbP}(\Omega) \otimes L^{2}(U)}
=
\left( \sum_{k \geq M+1} \lambda_k \right)^{\frac{1}{2}}.
\]
\end{theorem}

\begin{remark}
We conclude that the infimum above is achieved when $H_M =
\mbox{span}
\{\phi_1,\dots,\phi_{M}\}$ i.e., for the truncated KL
expansion.
\end{remark}

\begin{remark}
The KL expansion is largely a theoretical tool for signal
analysis. The main difficulty in its construction arises in estimation
of the random variables $Y_1(\omega), \dots,, Y_M(\omega)$. Although
these are mutually uncorrelated, in general they are not independent,
leading to a high dimensional joint distribution estimation problem.
Even for moderate dimension $M$, the number of realizations of
$v(\bx,\omega)$ needed to construct \revt{the joint probability
distribution function (pdf)} becomes impractical. However, for the
purposes of detecting anomalous signals and building a classifier,
only the eigenpairs $\{\lambda_k,\phi_k\}_{k=1}^M$ are needed, a
significantly easier problem. This can be achieved by constructing the
covariance matrix from realizations of $v(\bx,\omega)$ and computing
the eigenvalues and eigenvectors (See the method of
snapshots, \cite{Castrillon2002}).
\end{remark}

\subsection{Approach} In this section we show how to construct
subspaces that allow good separation between the classes. Recall from
our introductory example that if a Fourier basis is chosen for the
representation of the signal, separation between the signal components
can be found, making it easier to classify.

Suppose $u^{\bA}$, which we will refer as the nominal signal
    $v(\bx,\omega) - E_v$, and $u^{\bB},u^{\bC}$ are random field
    signals that belong in the Bochner space
    $L^{2}_{\bbP}(\Omega;L^{2}(U))$. The key question is, can we find
    a suitable tensor basis in $L^{2}_{\bbP}(\Omega;L^{2}(U))$ that
    can reveal separation between the signals? (See
    Figure \ref{fig:separationspaces}(a)). This is achieved by using a
    KL expansion together with anomaly detection.

Our novel approach to machine learning classification centrally involves
anomaly detection: identification of signals defined on the domain $U$ that
do not belong to a currently designated `nominal' family of finite
dimensional truncated KL expansions $v_M(\bx,\omega) - E_v = \sum_{k =
  1}^{M} \lambda^{\frac{1}{2}}_{k}$ $\phi_k(\bx) Y_{k}(\omega).$ To be
more precise, we seek to detect signals orthogonal to the eigenspace
spanned by $P_0:=\{\phi_1,\dots,\phi_M\}$.

Suppose that $W = \Span{\xi_1,\dots,\xi_a} \subset P_0^{\perp}$ 
where $\{\xi_1,\dots,\xi_a\}$ form an orthonormal set and
\[
v(\bx,\omega) - E_v = \sum_{k=1}^{\infty} \lambda^{\frac{1}{2}}_{k}
\phi_k(\bx) Y_{k}(\omega).
\]
Since the basis functions $\{\xi_1,\dots,\xi_a\}$ are orthonormal,
the projection coefficients of the signal $v(\bx,\omega) - E_v$
for $i = 1,\dots,a$ are
\[
\begin{split}
\alpha_i(\omega) &= \int_{U} (v(\bx,\omega) - E_v) \xi_i\,\mbox{d} \bx
= 
\int_{U} \left(\sum_{k = 1}^{\infty} \lambda^{\frac{1}{2}}_{k} \phi_k(\bx) Y_{k}(\omega)\right)  \xi_i\,\mbox{d} \bx \\
&=
\int_{U} \left( \sum_{k = M+1}^{\infty} \lambda^{\frac{1}{2}}_{k} \phi_k(\bx) Y_{k}(\omega) \right)  \xi_i\,\mbox{d} \bx.
  \end{split}
\]
The last equality is due to the fact that $W \subset
P_0^{\perp}$. Using the Cauchy--Schwarz inequality it follows that
\[
\eset{\alpha_{i}(\omega)^2} = \sum_{k=M+1}^{\infty} \lambda_k.
\]
Thus for any orthonormal basis of $W$ the variance of the projection coefficients for the nominal signal $v(\bx,\omega) - E_v$ will depend on the small truncated eigenvalues of the KL expansion. The idea is that we want to pick a basis of $W$ such that projection coefficients are large if the signal is from class $u^\bB$ or $u^\bC$ and small if it is from the nominal signal class (See Figure \ref{fig:separationspaces} (b)).

\corb{We construct $W$  with a multilevel space that contains 
large components of the external;
  anomalous signals, for the purpose of rejecting them from the
  currently designated null/nominal class, resulting in improved
  classification.} Note that in fact the construction is elaborate and
  nontrivial.  It is based on differential operator-adapted
  multilevel methods from scientific computing and computational
  applied mathematics approaches for solving Partial Differential
  Equations (see \cite{DHeedene2005} and \cite{Castrillon2003}).

\begin{asum}
Without loss of generality assume that $E_v = 0$, and consider a
sequence of nested subspaces $P_0 \subset P_1 \dots \subset L^{2}(U)$
such that $\overline{\bigcup_{k \in \bbN_{0}} P_{k}} = L^{2}(U)$ and
$P_{0} := \spn \{\phi_1,\phi_2,$ $\dots,\phi_M \}$.  Furthermore, let
the subspaces $S_k \subset L^{2}(U)$, for $k = 0,1,2,\dots$, be
defined by $P_{k+1} = P_{k} \oplus S_{k}$, so that $\overline{ P_0
  \bigoplus_{k \in \bbN_{0}} S_{k}} = L^{2}(U).$
\label{mls:assum1}
\end{asum}

\begin{asum}
For all $l \in \bbN_0$ let $\{ \{ \psi^{l}_{k} \}_{k = 1}^{M_l} \}_{l
  \in \bbN_{0}}$ be a collection of orthonormal functions with $S_{l}
= \spn \{\psi_1^l, \dots,$  $\psi_{M_{l}}^l \}$ and $M_l := \dim S_l$.
\label{mls:assum2}
\end{asum}

\begin{remark}
In practice the basis functions for the finite dimensional spaces $P_n
= P_0 \oplus S_0 \oplus \dots S_{n-1}$ will be constructed by using a
series of local Singular Value Decompositions (SVDs).  The space $P_n$
is assumed to be formed from the span of $N$ characteristic functions,
where the maximum level $n$ will be determined algorithmically.  The
construction of the basis for these spaces is intricate and is
described in detail in \cite{Castrillon2022} and in
\cite{Castrillon2022b}.
\end{remark}

\begin{remark}
Since the basis of $\bigoplus_{k \in \bbN_{0}} S_{k}$ is orthonormal,
for any function $u \in L^{2}(U)$ the orthogonal projection
coefficient onto the function $\psi^{l}_{k} \in W_{l}$ is
\[
d^l_k := \int_U u \psi^l_k \,\mbox{d} \bx.    
\label{mls:eqn0}
\]
\end{remark}

Given that $d^l_k$ are orthogonal projection coefficients (from
$S_{k}$) of a novel signal $u(\bx,\omega) \in
L^{2}_{\bbP}(\Omega;L^{2}(U))$, they provide a mechanism to detect the
magnitude of the novel part of the signal orthogonal to eigenspace
$P_0$. In more colloquial terms, we desire to detect the components of
$u(\bx,\omega)$ via stochastic properties different from those of the
eigenspace.  Suppose that $u(\bx,\omega) = v(\bx,\omega) +
w(\bx,\omega)$ i.e., the signal $u(\bx,\omega)$ is formed from
components $v(\bx,\omega)$ and $w(\bx,\omega) \in P_0^{\perp}$.  The
goal then is to detect the component $w(\bx,\omega)$ orthogonal to
eigenspace $P_0$. Thus $v(\bx,\omega)$ can represent a signal from the
nominal class and $u(\bx,\omega)$ the second class. However, in
practice we can only build the eigenspace for the truncated KL
expansion $v_M(\bx,\omega)$.  The following Lemma is stated from
\cite{Castrillon2022} and provides a mechanism relating strengths of
the classes with their coefficient magnitudes.

\begin{figure}[t]
\begin{tabular}{c c}
\begin{tikzpicture}[scale = 1,every node/.style={scale=1}]>=latex']
  \begin{scope}[cm={-1,-1,1,0,(0,0)},x=3.85mm,z=-1cm]
    \draw[-latex] (0,0,0) -- (4,0,0) node[anchor=north east]{};
    \draw[-latex] (0,0,0) -- (0,4,0) node[below]{};
    \draw[-latex] (0,0,0) -- (0,0,3) node[right]{};
  \end{scope}
  \shade[ball color=blue!20!white,opacity=1] (-2,2,0) circle (1cm);
  \shade[ball color=blue!50!green!20!white,opacity=1] (1,1,-2) circle (1cm);
  \shade[ball color=blue!50!red!20!white,opacity=1] (0,0,0) circle (1cm);
  \node[] at (-2.2,2.5,0) {$u^{\bA}$};
  \node[] at (1.7,2.25,0) {$u^{\bB}$};
  \node[] at (-0.3,0.3,0) {$u^{\bC}$};
  \node[] at (3.5,2.25,-1) {$L^{2}_{\bbP}(\Omega;L^{2}(U))$}; 
\end{tikzpicture} 
&
\begin{tikzpicture}[scale = 0.7,every node/.style={scale=0.7}]>=latex']
  \shade[ball color=blue!20!white,opacity=1] (0,0,0) circle (1cm);
  \shade[ball color=blue!20!white,opacity=1] (-3,-3,-3) circle (1cm);
  \node[] at (-3,-3,-3) {$u^{\bC}$};
  \shade[ball color=blue!50!red!20!white,opacity=1] (3,3,3) circle (1cm);
  \shade[ball color=blue!50!yellow!20!white,opacity=0.4] (0,0,0) circle (2.5cm);
  \shade[ball color=blue!50!green!20!white,opacity=0.2] (0,0,0) circle (4cm);
  \node[] at (0,0,0) {$u^{\bA}$};
  \node[] at (3,3,3) {$u^{\bB}$};
  \node[] at (-1.2,1.2) {$W \subset P_0^{\perp}$};
  \node[] at (-0.4,0.4) {$P_0$};
  \node[] at (-2.3,2.3) {$L^2(U)$}; 
\end{tikzpicture} \\
    (a) & (b)
\end{tabular}
\caption{Class separation in Hilbert spaces. (a) Given the right basis for the Bochner space $L^{2}_{\bbP}(\Omega;L^{2}(U))$, it is possible to find a separation between the classes. (b) 
Construction of subspace $W \subset P_0^{\perp}$ with which external
anomalous signals $u^{\bB}$ can be detected.
} \label{fig:separationspaces}
\end{figure}

\begin{lemma}
Suppose that $v \in L^{2}_{\bbP}(\Omega;L^{2}(U))$ with KL expansion
\[
v(\bx,\omega) = \sum_{p \in \bbN} \lambda^{\frac{1}{2}}_{k}
\phi_p(\bx) Y_{k}(\omega).
\]
Then for all $l \in \bbN_0$, $k = \{1,\dots, M_l\}$ and projection
coefficients
\[
d^l_k(\omega) = \int_U v(\bx,\omega) \psi^l_k \,\mbox{\emph{d}} \bx
\]
we have that a.s.
\[
\eset{d^{l}_k} =  0 
\ \ \,\mbox{\rm and}\,\,\ \ \ 
\eset{(d^{l}_k)^2} \leq \sum_{i\geq M+1} \lambda_{i}.
\]
\label{mls:lemma1}
\end{lemma}


If $u(\bx,\omega) = v(\bx,\omega)$, i.e. the signal
  $u(\bx,\omega)$ belongs to the nominal class, then the variances of
  the coefficients $d^l_k$ are controlled by the number of KL
  coefficients $M$. We can then use this to prove:
  
\begin{theorem} Suppose that we formulate the following Hypothesis
  test:
\[
   H_0:u(\bx,\omega) = v(\bx,\omega) \ \ \  H_A:u(\bx,\omega) \neq
   v(\bx,\omega).
  \]
  Let $1 \geq \alpha \geq 0 $ be the significance level, so that under
  $H_0$:
  \[
  \bbP \left(|d^l_k(\omega)| \geq \alpha^{-\frac{1}{2}}  \left(\sum_{i \geq M + 1}\lambda_i \right)^{\frac{1}{2}}
      \right)
\leq \alpha
\]
\label{mls:theo3}
\end{theorem}
\begin{proof}
  The result follows from Lemma \ref{mls:lemma1} and Chebyshev inequality.
  \end{proof}

\begin{figure}[htb]
\begin{center}
  \begin{tikzpicture}   
    \begin{scope}[xshift = -9cm, font=\scriptsize,
     dotout/.style={circle,draw=blueish,fill=blueish, inner
       sep=0pt,minimum size=1.5mm},
     dotin/.style={circle,draw=darkorange,fill=darkorange, inner
       sep=0pt,minimum size=1.5mm}]
    \draw [-stealth] (-2.75, 0) -- (2.75, 0) node [below left]  {\normalsize };
    \draw [-stealth] (0,-2.75) -- (0,2.75) node [below left] {\normalsize };
    \node at (0,   -3.5) [] {(a)};
    \node at (-0.0066,   -0.2705) [dotin] {};
    \node at (-1.2249,   -1.2646     ) [dotin] {};
    \node at (1.6408,    0.2353) [dotin] {};
    \node at (0.3640,    1.0547) [dotin] {};
    \node at (-1.7070,    1.7853) [dotin] {};
    \node at (-0.5740,   -1.8433) [dotin] {};
    \node at (1.2360,    0.9966) [dotin] {};
    
\node at (0.7586,0.1541) [dotout] {}; 
\node at (-1.4727,-0.8712) [dotout] {}; 
\node at (-1.5060,1.9038) [dotout] {}; 
\node at (-1.2364,-1.8543) [dotout] {}; 
\node at (-1.4171,-0.6950) [dotout] {}; 
\node at (0.3402,1.8921) [dotout] {}; 
\node at (-1.7066,-0.5399) [dotout] {}; 
\node at (1.2893,-0.7634) [dotout] {}; 
\node at (0.8916,-1.5164) [dotout] {}; 
\node at (1.7034,1.6631) [dotout] {}; 
\node at (-0.0294,-1.4581) [dotout] {}; 
\node at (0.6195,-0.6715) [dotout] {}; 
\node at (1.5605,1.5899) [dotout] {}; 
   
\coordinate (O) at (4,0);
\coordinate (P) at (5.3,0);
\draw[->, >=latex, gray, line width=4 pt] (O) -- (P);
    \end{scope}

    \begin{scope}[font=\scriptsize,
     dotout/.style={circle,draw=blueish,fill=blueish, inner
       sep=0pt,minimum size=1.5mm},
     dotin/.style={circle,draw=darkorange,fill=darkorange, inner
       sep=0pt,minimum size=1.5mm}]          
    \filldraw[fill=green!40!blue!20!white!50!] (0,0) ellipse (2 and 2);
    \node at (-0.5,-0.5) [dotin] {};
    \node at (0.2,-0.3) [dotin] {};
    \node at (0.5,0.5) [dotin] {};
    \node at (-0.25,0.25) [dotin] {};
    \node at (0.25,0.25) [dotin] {};
    \node at (-0.35,-0.25) [dotin] {};
    
    \node at (1.5,2) [dotin] {};

    \node at (2,2) [dotout] {};
    \node at (2,1.5) [dotout] {};
    \node at (2.25,2) [dotout] {};
    \node at (2.25,2.25) [dotout] {};

    \node at (-2,1) [dotout] {};
    \node at (-2.25,1) [dotout] {};
    \node at (-2,1.25) [dotout] {};
    \node at (-2.35,2.35) [dotout] {};

    \node at (1,-2) [dotout] {};
    \node at (1.25,-2) [dotout] {};
    \node at (1,-2.25) [dotout] {};
    \node at (0.75,-2) [dotout] {};
    \node at (-0.25,0.5) [dotout] {};
    \node at (-0.25,-0.5) [dotout] {};

    \draw [-stealth] (-2.75, 0) -- (2.75, 0) node [below left]
          {\normalsize $d^l_{k_1}$};

   \draw [-stealth] (0,-2.75) -- (0,2.75) node [below left]
         {\normalsize $d^l_{k_2}$};
    \node at (0,   -3.5) [] {(b)};
    \end{scope}
    
\end{tikzpicture}
\end{center}
\caption{Illustrative example of the separation between the
    projection coefficients of the nominal class and large anomalous
    signals based on the coefficients $d^l_k$. (a) The orange (nominal
    class) and blue dots (signal anomaly of the alternative class)
    corresponds to the original data in the feature space. These
    observations points are mixed with each other, which makes it hard
    to build a decision surface. (b) After applying the \corb{MOS}
    filter, the orange dots correspond to coefficients $d^l_k$ that
    are subject to the null hypothesis $H_0$ (nominal class). Thus, by
    Theorem \ref{mls:theo3}, the coefficients are expected to center
    around the origin with high probability.  The larger the number of
    KL eigenfunctions (given by parameter $M$) used to build the
    multilevel basis, the more likely the concentration of the
    coefficients is to be around the origin.  Conversely, under the
    alternative hypothesis $H_A$ (signal anomaly) the coefficients
    $d^l_k$ (blue dots) are likely not to concentrate around
    zero. This facilitates the construction of a separating surface
    between the two classes.}
\label{Separation}
\end{figure}

\revt{This theorem is consequential for building machine learning features. The conclusion is that features with separability characteristics can be constructed from the original data. These features are constructed from the residual spaces (residual principal components) of the truncated KL expansion. This is in contrast to PCA features that are usually picked from the principal components:}

\[
\revt{\overbrace{\phi_1 \,\,\, \phi_2 \,\,\, \dots \,\,\, \phi_M}^{\mbox{PCA Features}} \,\,\, \underbrace{\phi_{M+1} 
\,\,\, \phi_{M+2} \dots }_{
{
\underbrace{S_0 \bigoplus S_1 \bigoplus S_2 \bigoplus \dots}
_{\mbox{KL Features}}}
}
}
\]

\rev{If $u(\bx,\omega) = v(\bx,\omega)$ (i.e. the nominal class) then
  under the null hypothesis $H_0$ from Theorem \ref{mls:theo3} the
  coefficients $d^l_k$ will concentrate around the origin with
  controllable probability.  Conversely, under the alternative
  hypothesis $H_A$ (signal anomaly) the coefficients $d^l_k$ (blue
  dots) are likely not to concentrate around zero (though there is an
  unlikely possibility that some of them could be small). This makes
  it easier to build a separation surface for the two classes (See
  Figure \ref{Separation}). We can now separate the coefficients for
  the two classes more cleanly with a decision surface such as a
  Support Vector Machine (SVM) optimization (See
  \cite{Cristianini2000}). In particular, it is well suited with a
  Radial Basis Function (RBF) kernel.}

\begin{remark} It is important to note that the hypothesis test
  for Theorem \ref{mls:theo3} does not require any extra knowledge
  such as independence or the distribution of the underlying signal.
  This is in contrast to traditional hypothesis tests. 
\end{remark}

\begin{figure}[htb]
\centering
\begin{tikzpicture}[scale = 0.95, every node/.style={scale=0.95}]>=latex']
    \node at (0.6,-3) {
    \begin{tabular}{c}
      \\ \corb{(Class \textbf{A})} \\
      \includegraphics[trim=340 250 765
        100,clip,scale=0.09]{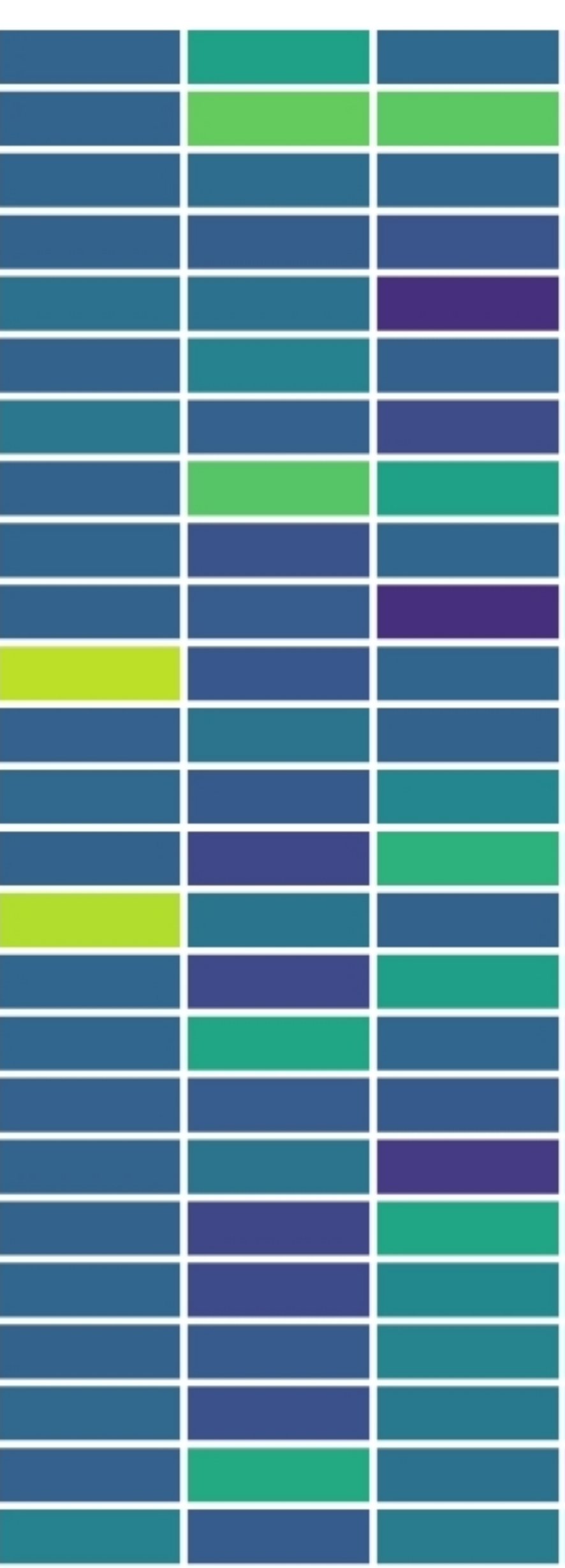}
    \end{tabular}
    };

    \node at (0.625,-5.5) {\small $\bm^{\bA}_1 \dots
      \bm^{\bA}_{m_1}$}; \node at (0.625,-10.5) {\small $\bm^{\bB}_1
      \dots \bm^{\bB}_{m_2}$};

    \node at (0.6,-8) {
    \begin{tabular}{c}
      \corb{} \\ \corb{(Class \textbf{B})}
      \\ \includegraphics[trim=905 250 200
        100,clip,scale=0.09]{./Figures/HeatMap.pdf}
    \end{tabular}
    };

    \begin{scope}[shift={(-0.38,0)}]

                  \foreach \y in{3} {
      \node[blockTwo] at (5.25,-\y-0.45) (nextblock\y)            
           {\begin{tabular}{{c}p{4.5cm}}
           \corb{Build \corb{MOS}} \\ 
               $ \bigoplus_{k \in \bbN_{0}} S_{k}$ \\
 \\
                 \corb{Feature Map (A)} \\
               $\bx^{\bA}_i = \Phi(\bm^{\bA}_{N_T+1},\dots, \bm^{\bA}_{m_1})$  \\
               $\by^{\bA}_i = 0$ \\
           \end{tabular}};
           \draw[vecArrowTwo] (1.65,-3.45) -- (nextblock\y.west);
           
                      
            }

                        \foreach \y in{4} {
      \node[blockTwo] at (5.25,-\y-4.45) (nextblock\y)            
           {\begin{tabular}{{c}p{8.5cm}}
            \corb{Feature Map (B)} \\ \\
               \corb{MOS} from (A) \\
               $\bigoplus_{k \in \bbN_{0}} S_{k}$ \\
               \\
               $\bx^{\bB}_i = \Phi(\bm^{\bB}_1,\bm^{\bB}_2,\dots, \bm^{\bB}_{m_2})$ \\
               $\by^{\bB}_i = 1$ \\
           \end{tabular}};
           \draw[vecArrowTwo] (1.65,-8.45) -- (nextblock\y.west);
           
          \node[blockTwo] at (12,-\y-2) (fourblock\y)            
           {\begin{tabular}{c}
               \corb{Support Vector Machine} \\
               \corb{Classifier} \\
           Input: Training Data \\
           $\{(\bx^{\bA}, \by^{\bA}),(\bx^{\bB}, \by^{\bB})
           \}$ \\
           \\
           ${\bw \in \R^{n}} 
           \frac{1}{n} \sum_{j = 1} \tau_j + \lambda \bw^T \bK \bw$ \\
           $y_j\left( \sum_{i=1}^{n} w_j K(\bx_i,\bx_j) + b \right) \geq 1 - \tau_j$, \\
           $\tau_j \geq 0$ \\
           \end{tabular}};

          \draw[vecArrowTwo] (nextblock\y.east) -- (fourblock\y.west);
                      
                        }

          \draw[vecArrowTwo] (nextblock3.east) -- (fourblock4.west);
          \draw[vecArrowTwo] (nextblock3.south) -- (nextblock4.north);                
       
\end{scope}         
\end{tikzpicture}
\caption{\corb{MOS-KL} training framework for binary
  classification using SVM. With a slight abuse of notation the map
  $\Phi: L^{2}(U) \rightarrow \bigoplus_{k \in \bbN_{0}} S_{k}$
  corresponds to the transformation of the signal $u(\bx,\omega)$ into
  the spaces $\bigoplus_{k \in \bbN_{0}} S_{k}$ and so provides the
  projection coefficients. The MOS are built from the
  classes where more data is available, in this case from the data of
  class $\bA$; $N_T < m_1$ samples are chosen ( $\bm^{\bA}_1,\dots,
  \bm^{\bA}_{N_T}$ ) to estimate the covariance function (matrix) and
  thus the $M$ eigenvalues and eigenfunctions.  The multilevel filter
  for $\bigoplus_{k \in \bbN_{0}} S_{k}$ is built from these
  eigenfunctions and the map $\Phi$ is applied to the data
  $\bm^{\bA}_{N+1},\dots, \bm^{\bA}_{m_1}$ and $\bm^{\bB}_1,,\dots,
  \bm^{\bB}_{m_2}$, and the SVM classifier is trained.}
\label{MultilevelSVM}
\end{figure}

The separations between signals depend on several factors: i) the
number of eigenfunctions $M$; ii) the accuracy of the computation of
the eigenspace (dependent on availability of data); iii) The presence
of noise in both signals.  In many practical applications such as for
gene expression data, $p$ will be large and $m$ relatively
small. Thus, generally, if we extract $N_T$ samples from class $\bA$
to construct the \corb{MOS} filter, there is no guarantee that
applying this filter to the remainder of the data we will obtain
near-zero values for coefficients. However, in general it is expected
that the multilevel coefficients for class $\bA$ will be smaller than
those for class $\bB$ due to Theorem \ref{mls:theo3}.  In
Figure \ref{MultilevelSVM} the classification training framework with
respect to two classes of data is shown. Note that this approach is
general and can additionally apply to data from more novel sources
arising from complex topologies.

We will test our algorithm on the ADNI dataset. Data used in the
preparation of this article were obtained from the Alzheimer’s Disease
Neuroimaging Initiative (ADNI) database
(\href{http://adni.loni.usc.edu}{adni.loni.usc.edu}). The ADNI was
launched in 2003 as a public-private partnership, led by Principal
Investigator Michael W. Weiner, MD. The primary goal of ADNI has been
to test whether serial magnetic resonance imaging (MRI), positron
emission tomography (PET), other biological markers, and clinical and
neuropsychological assessment can be combined to measure the
progression of mild cognitive impairment (MCI) and early Alzheimer’s
disease (AD).


\begin{table*}[t]
\centering
\small
\setlength{\tabcolsep}{4.5pt}
\renewcommand{\arraystretch}{1.15}
\begin{tabular}{@{}l l >{\columncolor{mlsbg}}c c >{\columncolor{mlsbg}}c c >{\columncolor{mlsbg}}c c >{\columncolor{mlsbg}}c c c@{}}
\toprule
& & \multicolumn{2}{c}{SVM} & \multicolumn{2}{c}{MLP} & \multicolumn{2}{c}{ResNet-3} & \multicolumn{2}{c}{ResNet-30} & \\
\cmidrule(lr){3-4}\cmidrule(lr){5-6}\cmidrule(lr){7-8}\cmidrule(lr){9-10}
\multicolumn{1}{c}{Task} & Metric & MOS-KL & Orig. & MOS-KL & Orig. & MOS-KL & Orig. & MOS-KL & Orig. & \multirow{-2}{*}{Baseline\textsuperscript{a}} \\
\midrule
\multirow{4}{*}{AD vs.\ CN} & Accuracy & 0.840\textsuperscript{L0} & 0.857 & 0.831\textsuperscript{L0} & \textbf{0.876} & 0.776\textsuperscript{L3} & 0.832 & 0.576\textsuperscript{L3} & 0.623 & 0.853\,\textsuperscript{RF} \\
 & Sensitivity & 0.802 & 0.878 & 0.778 & 0.871 & 0.760 & 0.869 & 0.538 & 0.778 & \textbf{0.927} \\
 & Specificity & 0.878 & 0.836 & \textbf{0.884} & 0.880 & 0.791 & 0.796 & 0.613 & 0.469 & 0.780 \\
 & AUC & 0.920 & 0.947 & 0.890 & 0.945 & 0.831 & 0.913 & 0.586 & 0.700 & \textbf{0.955} \\
\midrule
\multirow{4}{*}{AD vs.\ MCI} & Accuracy & \textbf{0.883}\textsuperscript{L3} & 0.834 & 0.845\textsuperscript{L3} & 0.837 & 0.817\textsuperscript{L1} & 0.788 & 0.632\textsuperscript{L0} & 0.633 & 0.837\,\textsuperscript{RUS} \\
 & Sensitivity & \textbf{0.856} & 0.689 & 0.849 & 0.730 & 0.796 & 0.618 & 0.636 & 0.393 & 0.704 \\
 & Specificity & 0.909 & \textbf{0.979} & 0.841 & 0.944 & 0.838 & 0.958 & 0.629 & 0.872 & 0.971 \\
 & AUC & \textbf{0.946} & 0.944 & 0.911 & 0.915 & 0.884 & 0.883 & 0.675 & 0.682 & 0.904 \\
\midrule
\multirow{4}{*}{CN vs.\ MCI} & Accuracy & \textbf{0.963}\textsuperscript{L3} & 0.910 & 0.941\textsuperscript{L3} & 0.874 & 0.884\textsuperscript{L2} & 0.785 & 0.649\textsuperscript{L1} & 0.610 & 0.913\,\textsuperscript{RUS} \\
 & Sensitivity & 0.948 & \textbf{0.986} & 0.914 & 0.974 & 0.844 & 0.981 & 0.629 & 0.941 & 0.839 \\
 & Specificity & 0.977 & 0.834 & 0.968 & 0.774 & 0.925 & 0.588 & 0.668 & 0.279 & \textbf{0.988} \\
 & AUC & \textbf{0.984} & \textbf{0.984} & 0.979 & 0.966 & 0.950 & 0.949 & 0.713 & 0.726 & 0.775 \\
\bottomrule
\end{tabular}
\caption{ADNI plasma (M12) --- radial-kernel SVM, $M=5$, leave-10-pairs-out cross-validation. Shaded columns use the MOS-KL representation; \emph{Orig.} is the same classifier on the original features. Bold marks the best value in each row.}
\label{tab:adni}
\smallskip
{\footnotesize\raggedright
\textsuperscript{a}\,Strongest of the three non-MOS-KL ensemble baselines, selected by accuracy and then reported on all four metrics: \textsuperscript{RF} random forest, \textsuperscript{GB} gradient boosting, \textsuperscript{RUS} RUSBoost.\par
Each MOS-KL accuracy carries a superscript L$\ell$ giving the level that maximises that classifier's accuracy; all four metrics for the classifier are reported at that level, and levels are chosen per classifier and are not comparable across columns.\par
Sensitivity and specificity must be read as a pair: the highest sensitivity is usually attained by a majority-biased model whose specificity is correspondingly low.\par}
\end{table*} 

\begin{table*}[t]
\centering
\small
\setlength{\tabcolsep}{4.5pt}
\renewcommand{\arraystretch}{1.15}
\begin{tabular}{@{}l l >{\columncolor{mlsbg}}c c >{\columncolor{mlsbg}}c c >{\columncolor{mlsbg}}c c >{\columncolor{mlsbg}}c c c@{}}
\toprule
& & \multicolumn{2}{c}{SVM} & \multicolumn{2}{c}{MLP} & \multicolumn{2}{c}{ResNet-3} & \multicolumn{2}{c}{ResNet-30} & \\
\cmidrule(lr){3-4}\cmidrule(lr){5-6}\cmidrule(lr){7-8}\cmidrule(lr){9-10}
\multicolumn{1}{c}{Task} & Metric & MOS-KL & Orig. & MOS-KL & Orig. & MOS-KL & Orig. & MOS-KL & Orig. & \multirow{-2}{*}{Baseline\textsuperscript{a}} \\
\midrule
\multirow{4}{*}{AD vs.\ CN} & Accuracy & 0.874\textsuperscript{L7} & 0.875 & 0.849\textsuperscript{L8} & 0.838 & 0.844\textsuperscript{L0} & 0.802 & 0.683\textsuperscript{L0} & 0.624 & \textbf{0.881}\,\textsuperscript{GB} \\
 & Sensitivity & \textbf{0.881} & 0.855 & 0.852 & 0.822 & 0.842 & 0.789 & 0.684 & 0.602 & 0.824 \\
 & Specificity & 0.867 & 0.895 & 0.845 & 0.855 & 0.845 & 0.815 & 0.682 & 0.647 & \textbf{0.939} \\
 & AUC & 0.934 & 0.942 & 0.915 & 0.914 & 0.911 & 0.882 & 0.739 & 0.669 & \textbf{0.943} \\
\midrule
\multirow{4}{*}{AD vs.\ EMCI} & Accuracy & \textbf{0.870}\textsuperscript{L2} & 0.866 & 0.839\textsuperscript{L0} & 0.832 & 0.838\textsuperscript{L5} & 0.805 & 0.711\textsuperscript{L2} & 0.638 & 0.840\,\textsuperscript{GB} \\
 & Sensitivity & \textbf{0.870} & 0.831 & 0.862 & 0.815 & 0.868 & 0.786 & 0.725 & 0.578 & 0.783 \\
 & Specificity & 0.870 & \textbf{0.901} & 0.816 & 0.849 & 0.809 & 0.825 & 0.697 & 0.698 & 0.897 \\
 & AUC & 0.935 & \textbf{0.942} & 0.914 & 0.906 & 0.913 & 0.885 & 0.771 & 0.697 & 0.927 \\
\midrule
\multirow{4}{*}{AD vs.\ LMCI} & Accuracy & 0.611\textsuperscript{L7} & 0.609 & 0.619\textsuperscript{L6} & 0.604 & 0.610\textsuperscript{L4} & 0.586 & 0.558\textsuperscript{L3} & 0.503 & \textbf{0.629}\,\textsuperscript{RUS} \\
 & Sensitivity & 0.602 & 0.340 & \textbf{0.625} & 0.469 & 0.608 & 0.417 & 0.557 & 0.351 & 0.584 \\
 & Specificity & 0.621 & \textbf{0.878} & 0.613 & 0.739 & 0.612 & 0.756 & 0.559 & 0.655 & 0.674 \\
 & AUC & 0.642 & \textbf{0.664} & 0.660 & 0.658 & 0.652 & 0.624 & 0.577 & 0.505 & 0.662 \\
\midrule
\multirow{4}{*}{CN vs.\ EMCI} & Accuracy & 0.695\textsuperscript{L2} & \textbf{0.698} & 0.658\textsuperscript{L4} & 0.652 & 0.648\textsuperscript{L7} & 0.625 & 0.565\textsuperscript{L2} & 0.543 & 0.676\,\textsuperscript{GB} \\
 & Sensitivity & 0.699 & 0.735 & 0.660 & 0.656 & 0.648 & 0.642 & 0.545 & 0.560 & \textbf{0.763} \\
 & Specificity & \textbf{0.691} & 0.660 & 0.656 & 0.648 & 0.649 & 0.608 & 0.585 & 0.526 & 0.589 \\
 & AUC & 0.756 & \textbf{0.766} & 0.715 & 0.707 & 0.702 & 0.673 & 0.589 & 0.562 & 0.729 \\
\midrule
\multirow{4}{*}{CN vs.\ LMCI} & Accuracy & 0.718\textsuperscript{L0} & \textbf{0.726} & 0.698\textsuperscript{L4} & 0.694 & 0.692\textsuperscript{L7} & 0.654 & 0.563\textsuperscript{L2} & 0.539 & 0.711\,\textsuperscript{RUS} \\
 & Sensitivity & 0.714 & \textbf{0.801} & 0.684 & 0.745 & 0.671 & 0.719 & 0.550 & 0.623 & 0.713 \\
 & Specificity & \textbf{0.723} & 0.650 & 0.712 & 0.643 & 0.713 & 0.589 & 0.577 & 0.454 & 0.709 \\
 & AUC & 0.787 & \textbf{0.800} & 0.753 & 0.765 & 0.751 & 0.713 & 0.584 & 0.557 & 0.760 \\
\midrule
\multirow{4}{*}{EMCI vs.\ LMCI} & Accuracy & 0.689\textsuperscript{L8} & 0.671 & \textbf{0.706}\textsuperscript{L0} & 0.695 & 0.696\textsuperscript{L2} & 0.667 & 0.578\textsuperscript{L6} & 0.581 & 0.693\,\textsuperscript{GB} \\
 & Sensitivity & 0.671 & \textbf{0.739} & 0.682 & 0.722 & 0.670 & 0.690 & 0.573 & 0.633 & 0.732 \\
 & Specificity & 0.707 & 0.603 & \textbf{0.729} & 0.668 & 0.721 & 0.645 & 0.583 & 0.529 & 0.653 \\
 & AUC & 0.732 & 0.739 & 0.755 & 0.761 & 0.748 & 0.739 & 0.600 & 0.622 & \textbf{0.769} \\
\bottomrule
\end{tabular}
\caption{CSF SOMAscan 7k --- linear-kernel SVM, $M=8$.}
\label{tab:csf}
\end{table*} 

\begin{table*}[t]
\centering
\small
\setlength{\tabcolsep}{4.5pt}
\renewcommand{\arraystretch}{1.15}
\begin{tabular}{@{}l l >{\columncolor{mlsbg}}c c >{\columncolor{mlsbg}}c c >{\columncolor{mlsbg}}c c >{\columncolor{mlsbg}}c c c@{}}
\toprule
& & \multicolumn{2}{c}{SVM} & \multicolumn{2}{c}{MLP} & \multicolumn{2}{c}{ResNet-3} & \multicolumn{2}{c}{ResNet-30} & \\
\cmidrule(lr){3-4}\cmidrule(lr){5-6}\cmidrule(lr){7-8}\cmidrule(lr){9-10}
\multicolumn{1}{c}{Task} & Metric & MOS-KL & Orig. & MOS-KL & Orig. & MOS-KL & Orig. & MOS-KL & Orig. & \multirow{-2}{*}{Baseline\textsuperscript{a}} \\
\midrule
\multirow{4}{*}{AD vs.\ CN} & Accuracy & 0.668\textsuperscript{L6} & 0.680 & \textbf{0.699}\textsuperscript{L2} & 0.687 & 0.684\textsuperscript{L4} & 0.684 & 0.570\textsuperscript{L2} & 0.600 & 0.678\,\textsuperscript{RUS} \\
 & Sensitivity & 0.688 & \textbf{0.771} & 0.697 & 0.723 & 0.668 & 0.727 & 0.563 & 0.665 & 0.700 \\
 & Specificity & 0.648 & 0.588 & \textbf{0.700} & 0.650 & 0.699 & 0.641 & 0.576 & 0.535 & 0.656 \\
 & AUC & 0.734 & \textbf{0.765} & 0.751 & 0.746 & 0.736 & 0.752 & 0.591 & 0.647 & 0.720 \\
\bottomrule
\end{tabular}
\caption{newAD cohort --- linear-kernel SVM, $M=8$.}
\label{tab:newad}
\end{table*} 

\section{Results}

We next test the multilevel features on three Alzheimer's disease--related
data sets.

\begin{itemize}

\item  \textbf{Blood plasma:} The first is drawn from the Alzheimer’s Disease Neuroimaging Initiative (ADNI), a longitudinal multicenter
study that was launched in 2003 designed to develop biomarkers for
detection and tracking of Alzheimer's disease, currently includes
ADNI1, ADNIGO, ADNI2, ADNI3, and ADNI4 cohort \cite{Petersen2010}.
This study primarily focuses on the ADNI1 cohort, which enrolled 209
AD, a combination of 742 Mild Cognitive Impairment (MCI) and Late Mild Cognitive Impairment (LMCI), and 112 Cognitive
Normal (CN) participants. For each participant, there is a visit code
of either baseline (BL), or one year later (M12) indicating the time
the plasma blood sample was collected. We performed tests on the M12
plasma proteomics dataset, which contains 146 proteins. Selecting only the M12 plasma proteomics dataset, the number of participants in each Alzheimer group is 54 CN, 96 AD, and 346 MCI/LMCI samples.  \jc{Blood plasma proteomic expressions are considered noisy and more difficult to detect AD due to the presence of contaminants from non-cerebral organs. Despite this, the results are highly accurate using MOS-KL features and an appropriate ML algorithm. Furthermore, a blood draw is non-invasive, which makes it attractive from a clinical perspective.} 
\item \textbf{CSF:} Second data set is the cerebrospinal fluid (CSF) proteome from ADNI,
profiled with the aptamer-based SomaScan 7k assay; after
$k$-nearest-neighbour imputation of the missing measurements ($k=5$), each
participant is characterized by the levels of 7006 proteins in the CSF,
spanning four diagnostic groups --- 167 CN, 182 Early Mild Cognitive
Impairment (EMCI), 221 Late Mild Cognitive Impairment (LMCI), and 138 AD
--- from which we form the six pairwise classification tasks (AD vs.\ CN,
AD vs.\ EMCI, AD vs.\ LMCI, CN vs.\ EMCI, CN vs.\ LMCI, and EMCI vs.\
LMCI). \jc{Due to the proximity to the brain, CSF is considered the gold standard for AD pathology detection. However, it requires a spinal tap, which is highly invasive and expensive.}

\item \textbf{Gene expression:} The third is the newAD data set \cite{Dileep2025}, which consists of gene expression
levels present in the DNA of a random sample of healthy and afflicted
participants: 184 healthy and 145 Alzheimer's disease--afflicted samples,
each described by 2053 gene expression levels, defining a single binary
healthy vs.\ AD classification task. 


\end{itemize}

We report binary classification results for all pairs of labels in the three
data sets. Taking AD versus CN in the ADNI blood plasma cohort as an example,
there are 96 AD and 54 CN samples. In every training--validation split the
covariance matrix for the AD class is estimated from the $96-54 = 42$ surplus AD
samples that remain after balancing the two classes, and the truncation
parameter is set to $M=5$. RBF SVMs are then trained on the balanced data using
leave-10-pairs-out cross-validation. For the CSF and newAD cohorts we set $M=8$
and use linear SVMs: their feature dimension is far larger ($7006$ and $2053$
features), and our experiments confirm that linear kernels generalize better
than RBF kernels in this high-dimensional, small-sample regime.

\begin{remark}
The leave-pair-out cross-validation is applied to both classes simultaneously
(pairwise). For two classes $\bA$ and $\bB$ with $N_{\bA}$ and $N_{\bB}$
samples, one sample is removed from each class as validation and the rest are
used for training; taking all combinations yields $N_{\bA}N_{\bB}$
training--validation tests. This estimator of the AUC is less biased than the
single leave-one-out scheme \cite{Airola2009}. For each split, the 42 surplus AD
samples used to build the covariance matrix are drawn at random, so that the
result is not biased toward any particular choice of covariance sample; the
remaining AD samples are matched to the CN samples, giving a class-balanced
training set and thereby correcting the class imbalance. For a faster
implementation we use leave-10-pairs-out cross-validation, which gives
$\bigl\lceil N_{\bA}/10\bigr\rceil \bigl\lceil N_{\bB}/10\bigr\rceil \approx
N_{\bA}N_{\bB}/100$ tests. The trends in the results are preserved when moving
from leave-1-pair-out to leave-10-pairs-out, confirming that the coarser scheme
does not distort the comparison.
\end{remark}

Across the three data sets a consistent pattern emerges: the multilevel (MOS-KL) features help most when the discriminative signal is hidden --- distributed across subtle combinations of features that are not individually informative --- and add little when that signal is already overt, either because the task is easy (a clear separation such as AD versus CN) or because the assay is information-rich (the $7006$-protein CSF panel). 
On the ADNI plasma cohort, where each subject is described by only 146 proteins and the disease signal is diffusely encoded across them rather than carried by any obvious marker, the RBF SVM on the normalized MOS-KL projection coefficients attains its highest performance on the CN versus LMCI task, with an accuracy of $96.3\%$ and an AUC of $0.984$; the same
SVM on the original features reaches only $91.0\%$ accuracy at the same AUC
($0.984$), so the multilevel representation yields a markedly better-calibrated
decision boundary on this balanced task (Table~\ref{tab:adni}). The advantage
persists on the other hard, disease-stage-adjacent contrast, AD versus LMCI,
where MOS-KL improves accuracy from $83.4\%$ to $88.3\%$; on both tasks the
MOS-KL SVM also significantly outperforms Gradient Boosting, RUS Boost, and
Random Forest trained on the original features. In contrast, on the easy AD
versus CN contrast the MOS-KL and original-feature SVMs are statistically
indistinguishable ($84.0\%$ vs.\ $85.7\%$).

The CSF and newAD results support the same hypothesis. The CSF SomaScan assay
already provides a rich, high-dimensional description of each subject, so little
additional predictive structure remains to be uncovered: MOS-KL matches the
original features on the easy AD versus CN task ($87.4\%$ vs.\ $87.5\%$, AUC
$0.93$) and gives only a marginal improvement on the hardest EMCI versus LMCI
contrast ($68.9\%$ vs.\ $67.1\%$). On the newAD gene-expression data the two
representations are likewise comparable ($66.8\%$ vs.\ $68.0\%$). We interpret this to mean that the multilevel decomposition surfaces combinations of features that jointly explain the AD outcome but are not individually apparent --- exactly what is needed when the discriminative information lies deep in the data rather than in any single marker (ADNI plasma). When the features are already highly informative (CSF) or the classes are easily separated (AD versus CN), the relevant signal is close to the surface, and there is little hidden structure left for the decomposition to recover.

Crucially, these gains are not specific to the SVM. The MOS-KL representation
also improves the neural-network classifiers, raising the accuracy of the MLP,
ResNet-3, and ResNet-30 in $23$ of $30$ task$\times$model comparisons. As with
the SVM, the improvement is largest on the tasks where the discriminative signal is most deeply hidden: on CN versus LMCI the multilevel features lift the MLP by $+6.7$ points ($87.4\%
\rightarrow 94.1\%$) and ResNet-3 by $+10.0$ points ($78.5\% \rightarrow
88.4\%$), and on CSF they raise ResNet-3 on AD versus CN from $80.2\%$ to
$84.4\%$. As with the SVM, the neural-network models benefit less from MOS-KL on the comparatively easy AD-versus-CN plasma task, for which the original features already provide sufficient class separation. Aside from this expected exception, the improvements observed across multiple neural-network architectures demonstrate that the benefit of the multilevel representation is not specific to a single classifier.

\begin{remark}
Because of the limited sample size, training a complex neural architecture is difficult. 
Our ResNet-3 and ResNet-30 classifiers stack, respectively, 3 and 30 fully-connected residual blocks of width 64 (Each block consists of two fully connected (FC) layers separated by a ReLU activation function, together with an identity skip connection), trained for 30 epochs with Adam optimizer. 
On the ADNI blood plasma inputs (146 features) this gives roughly $3.5\times 10^{4}$ trainable parameters for ResNet-3 and
$2.6\times 10^{5}$ for ResNet-30, both far exceeding the $\approx 150$ training samples per fold.
Given the limited sample size relative to the number of trainable parameters, there is insufficient data to train these neural networks reliably, particularly the deeper ResNet-30 architecture. Consequently, their cross-validated classification accuracies are lower than those achieved by the SVM.
Even so, as noted above, the MOS-KL representation still improves these networks relative to the original features, so the difficulty lies with the network complexity and data sufficiency rather than with the multilevel
features.
\end{remark}

\begin{figure}[t]
\centering
\footnotesize
\begin{tabular}{ccc}
    \includegraphics[width=0.32\textwidth]{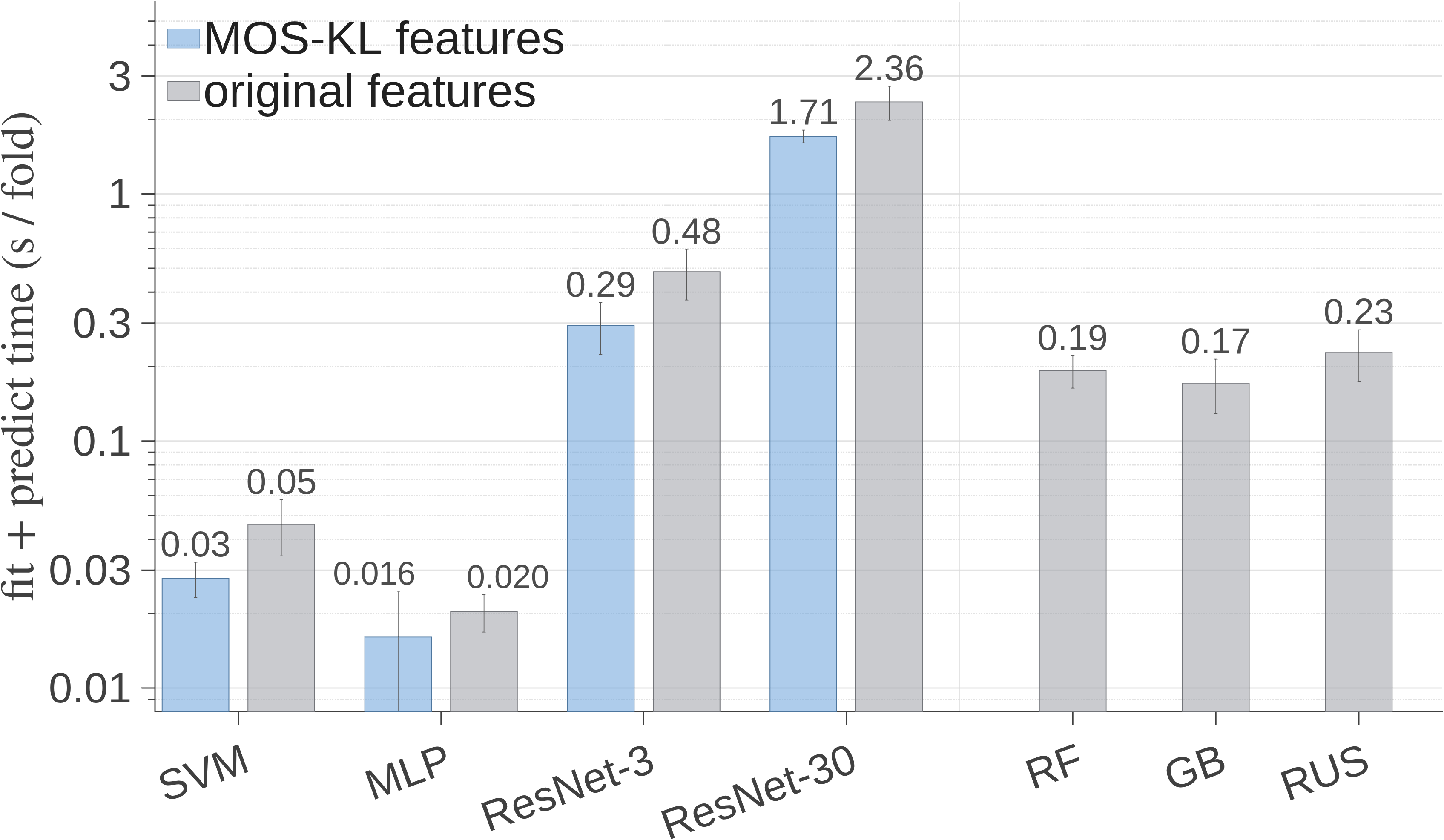}
    &
    \includegraphics[width=0.32\textwidth]{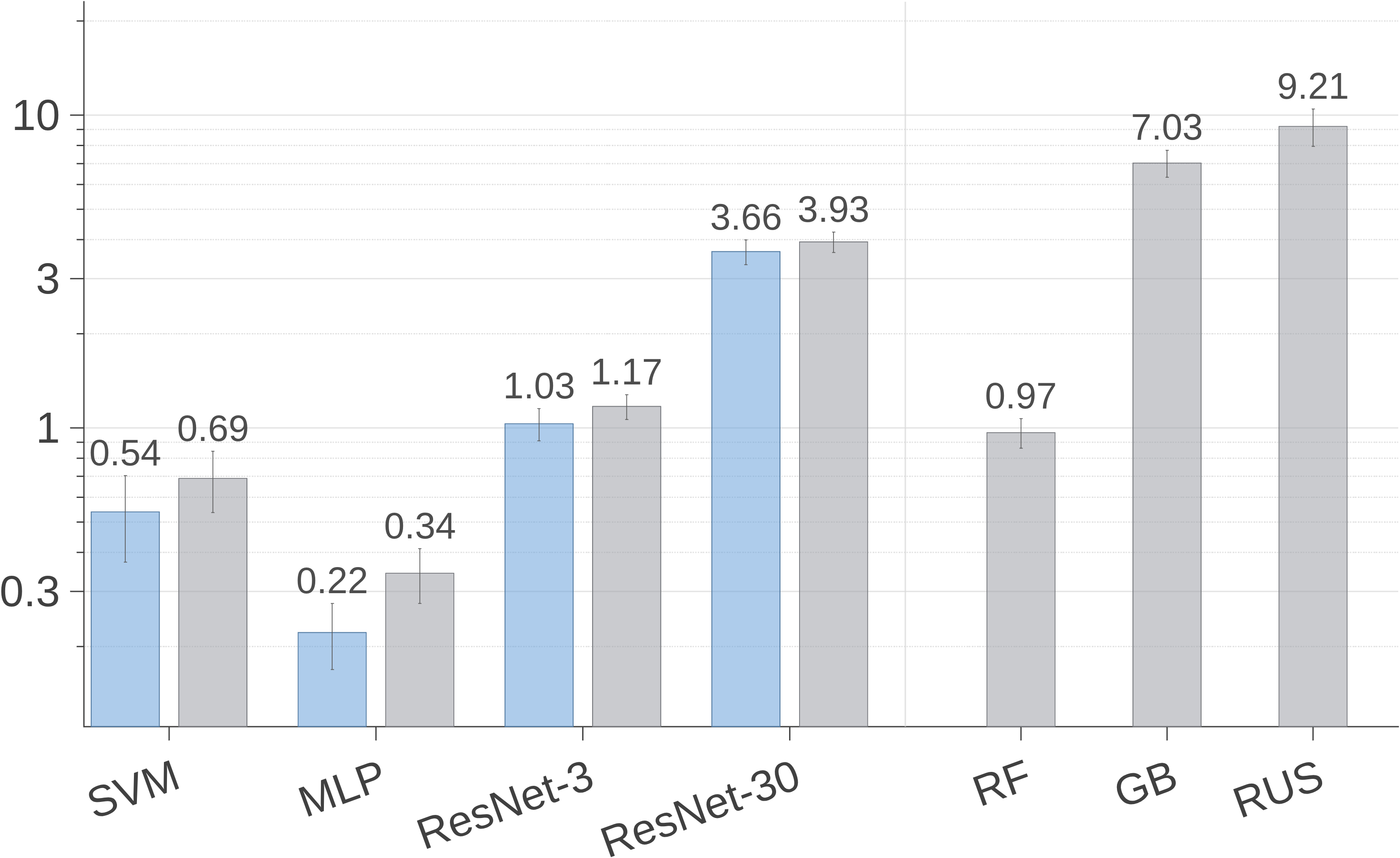}
    &
    \includegraphics[width=0.32\textwidth]{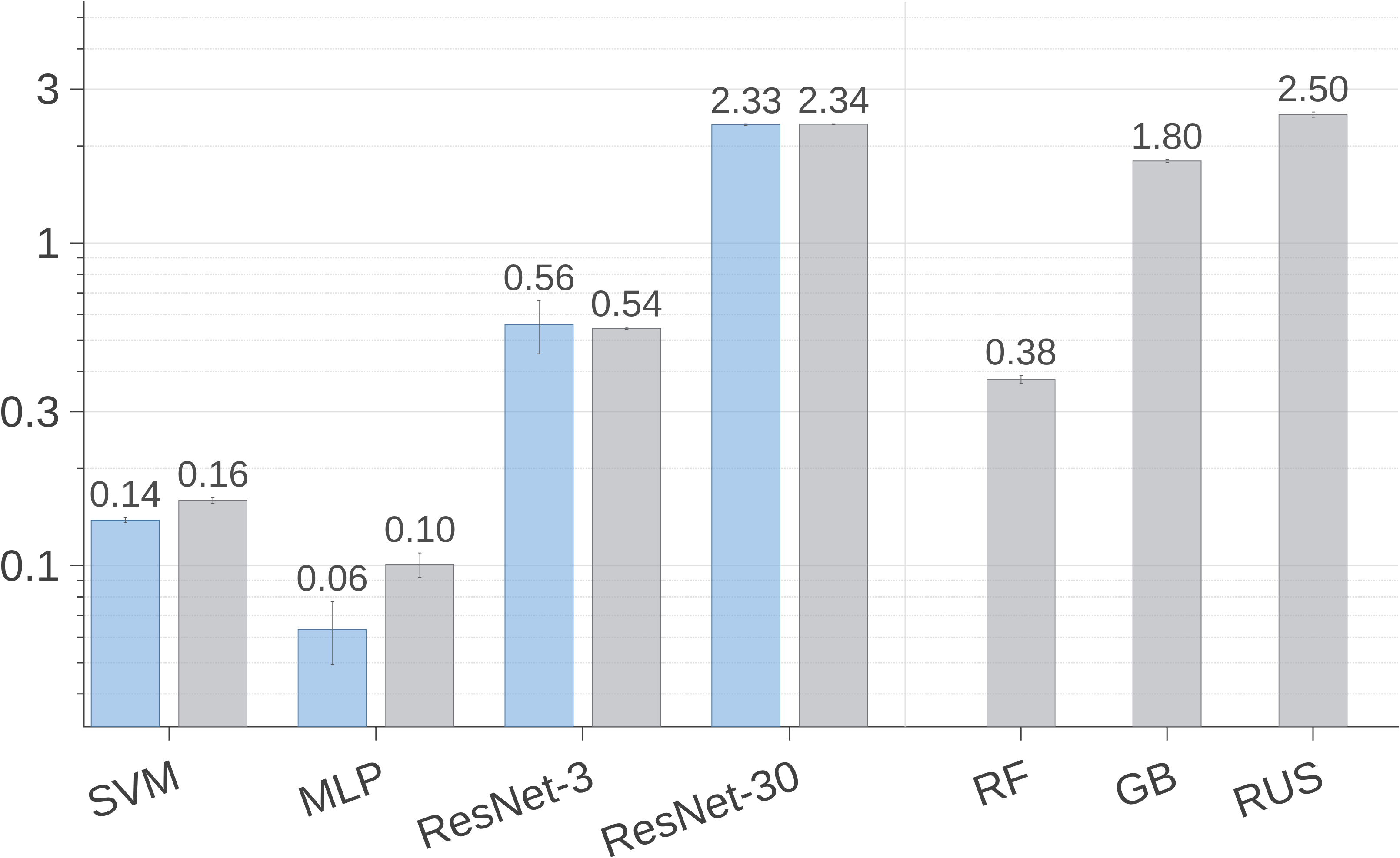}
    \\
    (a) ADNI
    &
    (b) CSF
    &
    (c) newAD
\end{tabular}
\caption{Runtime Comparison ADNI, CSF and newAD cohorts.
}
\label{fig:runtime_comparisons}
\end{figure}

\ks{Beyond improving classification accuracy, the multilevel representation also reduces computational cost through two complementary mechanisms. First, the MOS-KL projection replaces the original features with a smaller set of residual-space coefficients, moderately reducing the input dimension and the cost of each training iteration. This reduction alone, however, does not fully explain the observed speed-up because the decrease in dimension is relatively modest. More importantly, MOS-KL suppresses components dominated by within-class variability and makes the discriminative information more readily accessible to the classifier, thereby simplifying the learning problem and facilitating faster optimization, particularly for the MLP. Across the ADNI plasma, CSF, and newAD cohorts, MOS-KL reduces the mean per-fold fit-and-predict time by approximately 25\% for the SVM and 32\% for the MLP. On the ADNI plasma cohort specifically, the corresponding reductions are 40\% and 20\%, respectively. The computational advantage is particularly evident for the CSF cohort: although the SVM, MLP, and ensemble baselines achieve comparable classification accuracies, the MOS-KL SVM reduces the per-fold runtime by approximately 92\% relative to Gradient Boosting and 94\% relative to RUSBoost (Figure~\ref{fig:runtime_comparisons}). Thus, MOS-KL provides competitive predictive performance at a substantially lower computational cost.}

\section{Discussion}
In this paper we have introduced a novel approach for creating machine
learning features based on tensor product theory and stochastic
functional analysis.  The data are treated as random fields in a
Bochner space.  By constructing the appropriate spatial and stochastic
tensor bases a separation between the classes can be revealed and
constructed. This is achieved using a truncated KL expansion, combined
with construction of a basis for the subspace $W$ to detect nominal
signal projections in the anomalous subspace (complement of the
subspace spanned by the first $M$ principal components).  A multilevel
orthogonal basis is constructed to detect the magnitudes and locations
of these anomalies. Signals from different classes that are difficult
to distinguish in the original space are mapped to well-separated
coefficient magnitudes in the transformed space. An SVM classifier can
then more easily construct the separation boundary.  The performance
of the multilevel filter and the classifier depend on the availability
of a rich dataset for construction of the truncated eigenspace. For
signals that belong to a finite dimensional eigenspace and with
sufficient data it can be shown that our approach leads to perfect
classification (Theorem \ref{mls:theo3}). The performance increases
significantly as more data is available. This leads to a more accurate
covariance estimate, so that the separation between the classes
improves. This is confirmed from the numerical results obtained from
applying the multilevel filter on the semi-synthetic data created from
the GCM dataset. Furthermore, tests on the ADNI Alzheimer's Disease
proteomics dataset give rise to dramatic increases in accuracy.

\ks{A further advantage of is that MOS-KL features are not tied
to any particular classifier. The same projection coefficients that sharpen the
SVM also improve modern neural networks: across the ADNI plasma, CSF, and newAD
cohorts the MOS-KL representation increases the accuracy of the MLP, ResNet-3,
and ResNet-30 in 23 of 30 task$\times$model comparisons, with the largest gains on the tasks where the discriminative signal is most deeply hidden (for example, $+10.0$ accuracy points for
ResNet-3 and $+6.7$ for the MLP on CN versus LMCI). MOS-KL can therefore be
viewed as a model-agnostic feature-engineering layer that composes with
state-of-the-art learners rather than competing with them, while simultaneously
reducing the feature dimension --- and hence the training cost --- and
rebalancing the classes without wasting the data through its balanced covariance construction. Where the
discriminative signal is already easily accessible, such as the easy AD versus CN
contrast or the information-rich CSF assay, the transformation is at worst
neutral, so it can be adopted as a default preprocessing step with little
downside.}

The AUC values obtained from MOS-KL features are markedly higher than
benchmark results. Rehman et al. \cite{Rehman2024} reported models
incorporating a subset of ADNI blood plasma proteins panel together
with demographic and genetic variables (age, sex, education,
APOE4). In contrast, our use of the complete proteomic panel produced
substantially stronger performance than their Fig. 2, achieving an AUC
of 0.98 for MCI detection with a non-invasive blood test.


\ks{From a medical point of view, it is significant that the multilevel features help precisely where a test is most deployable.  Of the three modalities, blood plasma is by far the most accessible --- a plasma draw is minimally invasive, inexpensive, and already part of routine care --- 
whereas cerebrospinal fluid requires an invasive and painful lumbar puncture performed with a specialist, and gene-expression profiling is costly and labor-intensive. 
Yet it is on this low-dimensional plasma assay, and on the
clinically decisive early-stage contrasts, that MOS-KL delivers its largest gains, raising CN versus MCI accuracy to $96.3\%$ (AUC $0.984$) and AD versus MCI to $88.3\%$. Because the ADNI plasma samples can be collected at regular clinical visits, the MOS-KL representation
could turn a simple blood test into a sensitive screen for mild
cognitive impairment and Alzheimer's disease --- 
detecting the prodromal MCI stage, when disease-modifying therapies are most likely to be effective, and thereby opening a window for earlier diagnosis and intervention. In contrast,  CSF and genomic assays are more burdensome and cannot realistically provide early detection  at population scale. 
We recently obtained access to Bio-Hermes-001 \cite{biohermes2024}   and Framingham Heart Study Brain Aging Program (FHS-BAP) \cite{fhsbap_dataset} Prospectively, independent validation in external plasma-proteomic cohorts is a natural next step. }


There are still a number of strong avenues being explored based on the
work presented. In particular:
\begin{enumerate}[i)]

\item \rev{Extensions of the multilevel method to multi-class
  problems.}
 
    \item Effects of estimating the covariance structure on the
      accuracy of the classification.

    \item Optimal estimation of the parameters $M$ as well as the
      nested levels of the multilevel basis.
      
    \item Amelioration of the problem of overfitting. Since the
      multilevel filter leads to projection coefficients with greater
      distinguishability, our approach should be effective for this
      particular problem.

    \item \ks{Deeper integration of the multilevel features with modern     deep architectures. Our experiments already show that the multilevel representation improves multilayer perceptrons and residual networks; a promising direction is to learn the multilevel basis jointly with the network, or to use it as an input embedding for larger models, so that representation and classifier are optimized end-to-end.}

    \item Construction of optimal subspaces $W$ (not necessarily a
    multilevel construction) such that separations between classes are
    optimized.

    \item Exploration of connections between the stochastic
      transformations and high dimensional data described by tensors
      \cite{He2014,He2017a,He2017,Kour2023weighted,Kour2023}.

\end{enumerate}

\noindent \textbf{Acknowledgements:}
We acknowledge the assistance of Yulin Li and Hannah Pieper in
performing some of the machine learning numerical experiments.  In
addition, we are thankful to Tong Tong for helping to curate the ADNI
data. Furthermore, we thank Dileep Bhattacharya for providing to us
access the ADnew data from De novo analytics. This material is based upon work supported by the
National Science Foundation, Division of Mathematical Sciences, under
Grants No. 2347698 and 2319011.

Data collection and sharing for this project was funded by the
Alzheimer's Disease Neuroimaging Initiative (ADNI) (National
Institutes of Health Grant U01 AG024904) and DOD ADNI (Department of
Defense award number W81XWH-12-2-0012). ADNI is funded by the National
Institute on Aging, the National Institute of Biomedical Imaging and
Bioengineering, and through generous contributions from the following:
AbbVie, Alzheimer’s Association; Alzheimer’s Drug Discovery
Foundation; Araclon Biotech; BioClinica, Inc.; Biogen; Bristol-Myers
Squibb Company; CereSpir, Inc.; Cogstate; Eisai Inc.; Elan
Pharmaceuticals, Inc.; Eli Lilly and Company; EuroImmun;
F. Hoffmann-La Roche Ltd and its affiliated company Genentech, Inc.;
Fujirebio; GE Healthcare; IXICO Ltd.; Janssen Alzheimer Immunotherapy
Research \& Development, LLC.; Johnson \& Johnson Pharmaceutical
Research \& Development LLC.; Lumosity; Lundbeck; Merck \& Co., Inc.;
Meso Scale Diagnostics, LLC.; NeuroRx Research; Neurotrack
Technologies; Novartis Pharmaceuticals Corporation; Pfizer Inc.;
Piramal Imaging; Servier; Takeda Pharmaceutical Company; and
Transition Therapeutics. The Canadian Institutes of Health Research is
providing funds to support ADNI clinical sites in Canada. Private
sector contributions are facilitated by the Foundation for the
National Institutes of Health (\href{www.fnih.org}{www.fnih.org}). The
grantee organization is the Northern California Institute for Research
and Education, and the study is coordinated by the Alzheimer’s
Therapeutic Research Institute at the University of Southern
California. ADNI data are disseminated by the Laboratory for Neuro
Imaging at the University of Southern California.

  \newcommand{\beginsupplement}{%
        \setcounter{table}{0}
        \renewcommand{\thetable}{S\arabic{table}}%
        \setcounter{figure}{0}
        \renewcommand{\thefigure}{S\arabic{figure}}%
     }



\section*{Data, Materials, and Software Availability}


The Alzheimer’s Disease Neuroimaging
Initiative (ADNI) plasma proteomics data and CSF data are accessible through the
ADNI database (\url{https://adni.loni.usc.edu}) upon registration and
compliance with the ADNI Data Use Agreement.  The GCM gene expression
cancer dataset is included with our publicly available code. The newAD data is shared by \cite{Dileep2025} and we are not allowed to distribute. 

All MATLAB source code implementing the Multilevel Orthogonal Subspace
Karhunen–Loève (MOS–KL) feature framework, including data
preprocessing and classifier scripts, is openly available at our
public GitHub repository:
\url{https://github.com/SMLG-BU/MOS-KL}. 
The repository includes a detailed README file with instructions for
downloading the ADNI and CSF data, and preparing it for analysis. We use the
MIT License.

No new experimental data were generated for this work.  All analyses
were performed using existing, de-identified datasets and are fully
reproducible from the resources listed above.

\section*{Competing Interest Statement}
The authors declare no competing interests.

 \section*{PNAS SI Appendix: Performance tests}
\label{appendixA}

We test the performance of our MOS-KL features with data from the GCM
gene expression cancer dataset of \cite{Ramaswamy2001} (see also the
work from \cite{Tan2005}). The cancer data consist of 190 tumour ($m_1
= 190$ class $\bA$) and $m_2 = 90$ normal (class $\bB$) tissue data
with $p = 16,063$ gene expression levels. The domain $U$ is treated
one dimensional with $U:=[0,p-1]$ and the gene expression levels are
treated as one dimensional Haar functions on $U$.


To test the performance of the MOS-KL features accuracy-validation
tests were performed on semi-synthetic data that are created from this
dataset. The semi-synthetic data will allow us to study the
performance of the multilevel filter under different conditions. The
MOS-KL features with SVM RBF are compared with other popular ML
methods such as Gradient Boosting, RUS Boost and Random Forest with the
original features.


Semi-synthetic data is generated from this dataset to test performance
under varying conditions.  Our results show that the multilevel method
is particularly well suited, but not restricted, for extremely large
imbalanced datasets. Alternative approaches—such as
upsampling/downsampling, bootstrapping, and weighted classifiers—are
unsatisfactory, which has also motivated the development of
semi-supervised one class methods \cite{perera2021}.

  For each of these classes the covariance function and the mean are
    estimated with a method of snapshots by using all the available
    data. To generate the semi-synthetic data from the GCM dataset we
    can also use the KL expansion from Theorem \ref{remote:thm1}. This
    is a good choice as the realizations use the original covariance
    structure in the GCM dataset. In particular we have that

  \[
  \rev{
  \begin{split}
  \mbox{Cov}(v(\bx,\omega),v(\by,\omega)) & = \eset{
    \left(
    \sum_{k \in \bbN} \lambda^{\frac{1}{2}}_{k} \phi_k(\bx)
    Y_{k}(\omega) \right)
    \left(
    \sum_{l \in \bbN} \lambda^{\frac{1}{2}}_{l} \phi_l(\by)
    Y_{l}(\omega) \right) } \\
    & = 
  \sum_{k \in \bbN} \lambda_{k} \phi_k(\bx) \phi_k(\by).
  \end{split}
  }
  \]
  \rev{If $Y_k(\omega)$ for all $k\in \bbN$ are orthonormal in
    $L^{2}_{\bbP}(\Omega)$ then we have that $
    \mbox{Cov}(v(\bx,\omega),v(\by,\omega)) = \sum_{k \in \bbN}
    \lambda_{k} \phi_k(\bx) \phi_k(\by)$. This implies that we can
    replace $Y_k(\omega)$ for all $k \in \bbN$ with any set of zero
    mean, unit variance and orthogonal random variables $\tilde
    Y_k(\omega)$ and form the new random field
  \begin{equation}
    \tilde v(\bx,\omega)
    = E_{\tilde v} + \sum_{k \in \bbN} \lambda^{\frac{1}{2}}_{k}
  \phi_k(\bx) \tilde Y_{k}(\omega).
  \label{eqn:KLmod}
  \end{equation}
  It is easy to see that $\mbox{Cov}(\tilde v(\bx,\omega),\tilde
  v(\by,\omega)) = \mbox{Cov}(v(\bx,\omega),v(\by,\omega)) = \sum_{k
    \in \bbN} \lambda_{k} \phi_k(\bx) \phi_k(\by) $. Thus we can
  replace the model from equation \eqref{eqn:KL} in Theorem \ref{remote:thm1} with \eqref{eqn:KLmod} and $\tilde
  v(\bx,\omega)$ will have the same covariance structure as
  $v(\bx,\omega)$. Good choices for $\tilde Y_k(\omega)$ include
  assuming that they are Normal (or uniform). Thus
  $\tilde v(\bx,\omega)$ becomes a Gaussian process with the same
  covariance structure as $v(\bx,\omega)$.}

  \begin{remark} Note that KL expansion we use to construct the multilevel
    basis $P_0 \bigoplus_{k \in \bbN_{0}}$ $S_{k}$ from the
    semi-synthetic data and the KL expansion to generate the
    semi-synthetic data will not be the same. This process involves
    transforming the semi-synthetic data with a nonlinear sine
    function. This ensures that the transformed semi-synthetic data
    are not Gaussian processes and their KL expansions will be
    different from the KL expansions of original data.
    \end{remark}

  Initially the datasets are standardized to have zero mean and unit
  variance cross the features.  The KL expansion is applied to the
  covariance structures of both class $\bA$ and class $\bB$ data.  For
  example, for class $\bA$ using the truncated KL expansion
  realizations are generated from the eigenstructure of covariance
  function from class $\bA$:
\begin{equation}
  u^{\bA}_M(\bx,\omega) = E_{u^{\bA}} + \sum_{k=1}^{M_{\bA}}
  \sqrt{\lambda^{\bA}_k} \phi^{\bA}_k(\bx) Y^{\bA}_{k}(\omega).
\label{A:realization}
\end{equation}
It is assumed that the random field $u^{\bA}$ is a Gaussian process,
so that $Y^{\bA}_{k} \sim \mcN(0,1)$ are i.i.d. for $k =
1,\dots,M_{\bA}$.  The realizations are created by using a Gaussian
random number generator.  Similarly, the realizations from class $\bB$
are generated using the KL expansion:
\begin{equation}
u^{\bB}_M(\bx,\omega) = E_{u^{\bB}} + \sum_{k=1}^{M_{\bB}} \sqrt{\lambda^{\bB}_k}
\phi^{\bB}_k(\bx) Y^{\bB}_{k}(\omega).
\label{B:realization}
\end{equation}

We first test the ability of the multilevel filter to handle small to
large numbers of realizations with imbalanced datasets. Let $N_{\bA}$
be the number of realizations generated from the model
\eqref{A:realization}, and $\mcA_{N_{\bA}}$ the corresponding
dataset. Similarly we have defined $N_{\bB}$ and $\mcB_{N_{\bB}}$ for
the model \eqref{B:realization}. In the first experiment the dataset
$\mcA_{N_{\bA}}$ is generated with $M_{\bA} = 89$ terms in the KL
expansion. The dataset $\mcA_{N_{\bA}}$ is generated from the KL
expansion with $N_{\bA} =$ 150, 450, 1500, 10000 and 50000
realizations. These realizations are nested in the sense that the
random number generator seeds of the random variables
$\{Y^{\bA}_{1},\dots,Y^{\bA}_{M_{\bA}} \}$ are reset each time the
dataset $\mcA_{N_{\bA}}$ is created.  A single dataset
$\mcB_{N_{\bB}}$ is generated with $N_{\bB} = 100$ and $M_{\bB} = 89$
from model \eqref{B:realization}.

For each dataset $\mcA_{N_{\bA}}$, let $\mcA^{T}_{N_{\bA}}$ consist of
the first $N_{\bB}$ realizations and $\mcA^{C}_{N_{\bA } - N_{\bB}}$
be the rest of the data in $\mcA_{N_{\bA}}$ used to compute the
covariance matrix and construct the multilevel filter. The truncation
parameter for the KL expansion is set to $M = 39 $. The multilevel
filter built from the data in $\mcA^{C}_{N_{\bA}-N_{\bB}}$ is now
applied to the realizations in $\mcA^{T}_{N_{\bA}}$ (Class $\bA$) and
$\mcB_{N_{\bB}}$ (Class $\bB$). We obtain the datasets
$\mcA^{T,\mcM}_{N_{\bA}}$ and $\mcB_{N_{\bB}}^{\mcM}$, which are used
to train the SVM classifier for different nested $Levels$.

To test the accuracy of the multilevel SVM classifier we generate
validation datasets. For the class $\bA$, let $\mcA^{V}_{\tilde
  N_{\bA}}$ be the collection of $\tilde N_{\bA} = 10,000$ generated
realizations from the KL expansion in equation
\eqref{A:realization}. Conversely, the dataset $\mcB^{V}_{\tilde
  N_{\bB}}$ is generated for class $\bB$ with $\tilde N_{\bB} =
10,000$ from equation \eqref{B:realization}.  The multilevel filter is
then applied to the datasets $\mcA^{V}_{\tilde N_{\bA}}$ and
$\mcB^{V}_{\tilde N_{\bB}}$ and we obtain $\mcA^{V,\mcM}_{\tilde
  N_{\bA}}$ and $\mcB^{V,\mcM}_{\tilde N_{\bB}}$.

\begin{table*}[t]
\centering
\small
\setlength{\tabcolsep}{4.5pt}
\renewcommand{\arraystretch}{1.15}
\begin{tabular}{@{}l l >{\columncolor{mlsbg}}c c >{\columncolor{mlsbg}}c c >{\columncolor{mlsbg}}c c >{\columncolor{mlsbg}}c c c@{}}
\toprule
& & \multicolumn{2}{c}{SVM} & \multicolumn{2}{c}{MLP} & \multicolumn{2}{c}{ResNet-3} & \multicolumn{2}{c}{ResNet-30} & \\
\cmidrule(lr){3-4}\cmidrule(lr){5-6}\cmidrule(lr){7-8}\cmidrule(lr){9-10}
\multicolumn{1}{c}{Tumor} & Metric & MOS-KL & Orig & MOS-KL & Orig & MOS-KL & Orig & MOS-KL & Orig & \multirow{-2}{*}{Baseline\textsuperscript{a}} \\
\midrule
\multirow{4}{*}{150} & Accuracy & 0.898\textsuperscript{L0} & 0.897 & 0.899\textsuperscript{L4} & 0.929 & 0.895\textsuperscript{L1} & \textbf{0.948} & 0.791\textsuperscript{L0} & 0.699 & 0.868\,\textsuperscript{GB} \\
 & Sensitivity & 0.896 & 0.902 & 0.880 & 0.986 & 0.951 & 0.923 & 0.739 & 0.882 & \textbf{0.995} \\
 & Specificity & 0.900 & 0.892 & 0.919 & 0.873 & 0.839 & \textbf{0.974} & 0.843 & 0.515 & 0.742 \\
 & AUC & 0.964 & 0.963 & 0.960 & \textbf{0.992} & 0.974 & 0.989 & 0.873 & 0.822 & 0.990 \\
\midrule
\multirow{4}{*}{450} & Accuracy & 0.940\textsuperscript{L7} & 0.900 & 0.909\textsuperscript{L0} & 0.848 & 0.911\textsuperscript{L6} & \textbf{0.970} & 0.813\textsuperscript{L3} & 0.782 & 0.946\,\textsuperscript{RUS} \\
 & Sensitivity & 0.960 & 0.973 & 0.914 & \textbf{0.995} & 0.960 & 0.989 & 0.809 & 0.966 & 0.910 \\
 & Specificity & 0.921 & 0.827 & 0.904 & 0.702 & 0.862 & 0.952 & 0.817 & 0.598 & \textbf{0.982} \\
 & AUC & 0.987 & 0.979 & 0.970 & 0.990 & 0.974 & \textbf{0.997} & 0.880 & 0.930 & 0.897 \\
\midrule
\multirow{4}{*}{1,500} & Accuracy & 0.964\textsuperscript{L7} & 0.919 & 0.928\textsuperscript{L4} & 0.908 & 0.936\textsuperscript{L7} & \textbf{0.975} & 0.827\textsuperscript{L5} & 0.802 & 0.948\,\textsuperscript{RUS} \\
 & Sensitivity & 0.980 & 0.997 & 0.958 & \textbf{0.999} & 0.988 & \textbf{0.999} & 0.847 & 0.984 & 0.987 \\
 & Specificity & 0.947 & 0.842 & 0.898 & 0.818 & 0.885 & \textbf{0.950} & 0.807 & 0.621 & 0.910 \\
 & AUC & 0.994 & 0.983 & 0.970 & 0.994 & 0.977 & \textbf{1.000} & 0.895 & 0.949 & 0.994 \\
\midrule
\multirow{4}{*}{10,000} & Accuracy & \textbf{0.972}\textsuperscript{L6} & 0.927 & 0.917\textsuperscript{L2} & 0.842 & 0.928\textsuperscript{L6} & 0.916 & 0.826\textsuperscript{L7} & 0.867 & 0.952\,\textsuperscript{RUS} \\
 & Sensitivity & 0.986 & \textbf{1.000} & 0.948 & \textbf{1.000} & 0.983 & \textbf{1.000} & 0.785 & \textbf{1.000} & 0.993 \\
 & Specificity & \textbf{0.958} & 0.855 & 0.887 & 0.683 & 0.872 & 0.833 & 0.867 & 0.734 & 0.911 \\
 & AUC & 0.996 & 0.979 & 0.960 & 0.999 & 0.971 & \textbf{1.000} & 0.909 & 0.998 & 0.995 \\
\midrule
\multirow{4}{*}{50,000} & Accuracy & \textbf{0.973}\textsuperscript{L6} & 0.922 & 0.916\textsuperscript{L2} & 0.866 & 0.944\textsuperscript{L3} & 0.903 & 0.747\textsuperscript{L4} & 0.950 & 0.953\,\textsuperscript{RUS} \\
 & Sensitivity & 0.988 & \textbf{1.000} & 0.975 & \textbf{1.000} & 0.937 & \textbf{1.000} & 0.787 & \textbf{1.000} & 0.996 \\
 & Specificity & \textbf{0.959} & 0.845 & 0.857 & 0.732 & 0.950 & 0.805 & 0.707 & 0.900 & 0.910 \\
 & AUC & 0.996 & 0.975 & 0.963 & 0.999 & 0.987 & \textbf{1.000} & 0.815 & \textbf{1.000} & 0.996 \\
\bottomrule
\end{tabular}
\caption{GCM semisynthetic, $\sin(15 x)$ --- radial-kernel SVM, $M=39$, per-sample normalization, KL generation ($N_{\mathrm{KL}}=89$), 10{,}000 held-out test samples per class. Rows grouped by the number of generated Tumor training samples (Normal fixed at 100). Shaded columns use the MOS-KL representation; \emph{Orig.} is the same classifier on the original features. Bold marks the best value in each row.}
\label{tab:semisyn_sin15}
\smallskip
{\footnotesize\raggedright
\textsuperscript{a}\,Strongest of the three non-MOS-KL ensemble baselines, selected by accuracy and then reported on all four metrics: \textsuperscript{RF} random forest, \textsuperscript{GB} gradient boosting, \textsuperscript{RUS} RUSBoost.\par
Each MOS-KL accuracy carries a superscript L$\ell$ giving the level that maximises that classifier's accuracy; all four of its metrics are reported at that level, and levels are chosen per classifier and are not comparable across columns.\par
Sensitivity and specificity must be read as a pair: the highest sensitivity is usually attained by a majority-biased model whose specificity is correspondingly low.\par}
\end{table*} 

\begin{figure}[t]
  \centering
  \includegraphics[scale = 0.5]{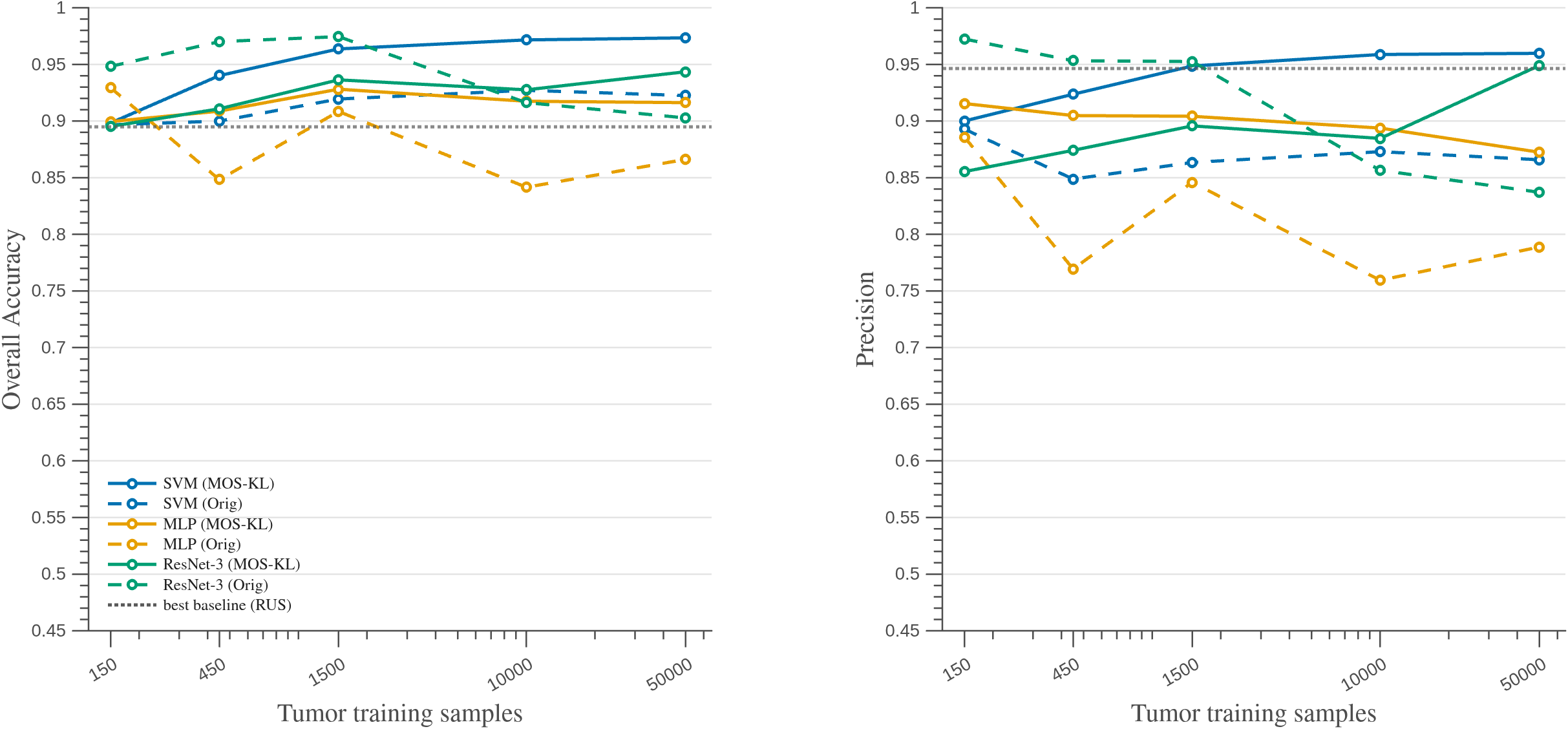}
  \caption{$\sin(15 x)$ --- Accuracy and precision comparison results as
    the number of samples in the training datasets increases for class
    $\bA$ $(\mcA_{{N_\bA}})$.}
  \label{fig:semisyn15}
\end{figure} 

\textbf{Test \#1:} We first apply the mildest deformation, updating every
realization $u^{\bA}_M(\bx,\omega_k)$ in the datasets $\mcA_{N_{\bA}}$ and
$\mcA^{V}_{\tilde N_{\bA}}$ as $u^{\bA}_M(\bx,\omega_k) \leftarrow \sin(15\,
u^{\bA}_M(\bx,\omega_k))$, and likewise every realization
$u^{\bB}_M(\bx,\omega_k)$ in $\mcB_{N_{\bB}}$ and $\mcB^{V}_{\tilde N_{\bB}}$ as
$u^{\bB}_M(\bx,\omega_k) \leftarrow \sin(15\, u^{\bB}_M(\bx,\omega_k))$. The
sine transform destroys the Gaussian structure of the realizations and mixes the
class information across scales, so the classes are no longer linearly separable
in the original coordinates. We sweep the number of generated Tumor training
samples over $N_{\bA} = 150,\,450,\,1500,\,10000$ and $50000$, with the Normal
class fixed at $N_{\bB} = 100$; the corresponding covariance sets
$\mcA^{C}_{N_{\bA}-N_{\bB}}$ therefore contain $50,\,350,\,1400,\,9900$ and
$49900$ realizations. For each configuration the multilevel filter is built with
truncation $M = 39$ over the nested spaces $S_{0}\oplus\dots\oplus S_{Level}$,
$Level = 0,1,\dots,7$, and the classifiers are evaluated on the held-out sets of
$10{,}000$ samples per class. We compare the multilevel (MOS-KL) features against
the original features for a radial SVM, an MLP, and two residual networks
(ResNet-3 and ResNet-30), and against the strongest of the Gradient Boosting, RUS
Boost, and Random Forest baselines on the original features; each MOS-KL entry is
reported at the level that maximizes that classifier's accuracy
(Table~\ref{tab:semisyn_sin15}).

\begin{remark}
  Recall that we use SVM with the fitPosterior option enabled in
  MATLAB \cite{Matlab2025,Platt2000,Tao2005}
\end{remark}

The central observation is that the MOS-KL + SVM RBF classifier \emph{improves as
more training realizations become available}: its accuracy rises monotonically
from $89.8\%$ at $N_{\bA}=150$ to $97.3\%$ at $N_{\bA}=50{,}000$, and the level
that achieves this optimum deepens from $Level = 0$ to $Level = 6$ as more data
sharpen the covariance estimate. The original-feature SVM does not benefit from
the extra data in the same way --- its accuracy stays near $90\%$
($89.7\%\rightarrow 92.2\%$) --- so the MOS-KL representation overtakes it and
attains the best SVM performance at every size beyond the smallest. The MOS-KL
SVM also consistently surpasses the strongest ensemble baseline (RUS Boost) at all sizes. 
At the smallest size the highly expressive ResNet-3 on the original features is momentarily competitive, but it does not scale: as the data gets more imbalanced, the multilevel SVM pulls clearly ahead. 
The multilevel features also help the neural networks --- MOS-KL
improves the MLP and ResNet-3 over their original-feature counterparts on the larger, more imbalanced sets --- confirming that the benefit is not tied to a single classifier, although SVM RBF with the multilevel features remains the strongest combination here.

Figure~\ref{fig:semisyn15} plots the overall accuracy (left) and the precision
(right) as functions of $N_{\bA}$, for MOS-KL + SVM, MOS-KL + MLP, MOS-KL + ResNet-3 and
their original-feature counterparts, each at its best-accuracy level; the dotted
horizontal line marks the strongest ensemble baseline. Both the accuracy and the
precision of the multilevel classifiers increase steadily (precision of MLP with MOS-KL feature slighlt declines but still significantly outperforms MLP with original feature) as the Tumor class
grows and the data become more imbalanced, 
whereas the original-feature models flatten or decline. 
The multilevel SVM is the only method whose accuracy and precision curves rise above the baseline and keep climbing.

\begin{remark}
  Although we use SVM RBF with the posterior option switched on, it is not the
  SVM itself that is robust to the imbalance. Recall that we remove the surplus
  Tumor training realizations in $\mcA^{C}_{N_{\bA}-N_{\bB}}$ and use them only to
  construct the covariance matrix, so the data that actually train the classifier
  are class-balanced. This is what makes the approach improve, rather than
  degrade, as the raw datasets become more imbalanced.
\end{remark}

\begin{table*}[t]
\centering
\small
\setlength{\tabcolsep}{4.5pt}
\renewcommand{\arraystretch}{1.15}
\begin{tabular}{@{}l l >{\columncolor{mlsbg}}c c >{\columncolor{mlsbg}}c c >{\columncolor{mlsbg}}c c >{\columncolor{mlsbg}}c c c@{}}
\toprule
& & \multicolumn{2}{c}{SVM} & \multicolumn{2}{c}{MLP} & \multicolumn{2}{c}{ResNet-3} & \multicolumn{2}{c}{ResNet-30} & \\
\cmidrule(lr){3-4}\cmidrule(lr){5-6}\cmidrule(lr){7-8}\cmidrule(lr){9-10}
\multicolumn{1}{c}{Tumor} & Metric & MOS-KL & Orig & MOS-KL & Orig & MOS-KL & Orig & MOS-KL & Orig & \multirow{-2}{*}{Baseline\textsuperscript{a}} \\
\midrule
\multirow{4}{*}{150} & Accuracy & 0.835\textsuperscript{L4} & 0.852 & 0.829\textsuperscript{L3} & 0.800 & 0.832\textsuperscript{L1} & \textbf{0.868} & 0.735\textsuperscript{L7} & 0.659 & 0.802\,\textsuperscript{GB} \\
 & Sensitivity & 0.824 & 0.849 & 0.764 & 0.904 & 0.893 & 0.904 & 0.721 & 0.834 & \textbf{0.954} \\
 & Specificity & 0.846 & 0.856 & \textbf{0.893} & 0.696 & 0.770 & 0.832 & 0.748 & 0.483 & 0.650 \\
 & AUC & 0.907 & 0.918 & 0.904 & 0.916 & 0.917 & 0.943 & 0.799 & 0.763 & \textbf{0.951} \\
\midrule
\multirow{4}{*}{450} & Accuracy & 0.906\textsuperscript{L7} & 0.859 & 0.845\textsuperscript{L5} & 0.786 & 0.849\textsuperscript{L3} & \textbf{0.925} & 0.722\textsuperscript{L1} & 0.694 & 0.903\,\textsuperscript{RUS} \\
 & Sensitivity & 0.926 & 0.958 & 0.907 & \textbf{0.994} & 0.772 & 0.986 & 0.677 & 0.966 & 0.945 \\
 & Specificity & 0.887 & 0.760 & 0.783 & 0.579 & \textbf{0.926} & 0.863 & 0.767 & 0.422 & 0.861 \\
 & AUC & 0.968 & 0.959 & 0.921 & 0.967 & 0.943 & \textbf{0.989} & 0.796 & 0.879 & 0.953 \\
\midrule
\multirow{4}{*}{1,500} & Accuracy & \textbf{0.936}\textsuperscript{L7} & 0.884 & 0.843\textsuperscript{L5} & 0.906 & 0.842\textsuperscript{L7} & 0.856 & 0.676\textsuperscript{L5} & 0.639 & 0.913\,\textsuperscript{RUS} \\
 & Sensitivity & 0.963 & 0.993 & 0.844 & 0.992 & 0.929 & \textbf{0.999} & 0.728 & \textbf{0.999} & 0.962 \\
 & Specificity & \textbf{0.909} & 0.775 & 0.842 & 0.821 & 0.754 & 0.712 & 0.623 & 0.278 & 0.865 \\
 & AUC & 0.979 & 0.960 & 0.902 & 0.986 & 0.910 & \textbf{0.993} & 0.733 & 0.952 & 0.976 \\
\midrule
\multirow{4}{*}{10,000} & Accuracy & \textbf{0.942}\textsuperscript{L6} & 0.871 & 0.848\textsuperscript{L4} & 0.873 & 0.843\textsuperscript{L6} & 0.722 & 0.745\textsuperscript{L2} & 0.805 & 0.910\,\textsuperscript{RUS} \\
 & Sensitivity & 0.982 & 0.999 & 0.939 & \textbf{1.000} & 0.920 & \textbf{1.000} & 0.737 & \textbf{1.000} & 0.974 \\
 & Specificity & \textbf{0.901} & 0.742 & 0.757 & 0.746 & 0.767 & 0.444 & 0.752 & 0.611 & 0.847 \\
 & AUC & 0.982 & 0.950 & 0.903 & 0.998 & 0.890 & \textbf{1.000} & 0.827 & 0.986 & 0.977 \\
\midrule
\multirow{4}{*}{50,000} & Accuracy & \textbf{0.945}\textsuperscript{L7} & 0.853 & 0.804\textsuperscript{L2} & 0.819 & 0.856\textsuperscript{L5} & 0.813 & 0.680\textsuperscript{L3} & 0.565 & 0.912\,\textsuperscript{RUS} \\
 & Sensitivity & 0.974 & \textbf{1.000} & 0.777 & \textbf{1.000} & 0.894 & \textbf{1.000} & 0.770 & \textbf{1.000} & 0.971 \\
 & Specificity & \textbf{0.917} & 0.705 & 0.626 & 0.060 & 0.819 & 0.625 & 0.591 & 0.129 & 0.854 \\
 & AUC & 0.983 & 0.944 & 0.879 & 0.993 & 0.924 & \textbf{0.997} & 0.730 & 0.786 & 0.977 \\
\bottomrule
\end{tabular}
\caption{GCM semisynthetic, $\sin(18 x)$ --- same setup as Table~\ref{tab:semisyn_sin15}.}
\label{tab:semisyn_sin18}
\end{table*} 

\begin{figure}[h]
  \centering
  \includegraphics[scale = 0.5]{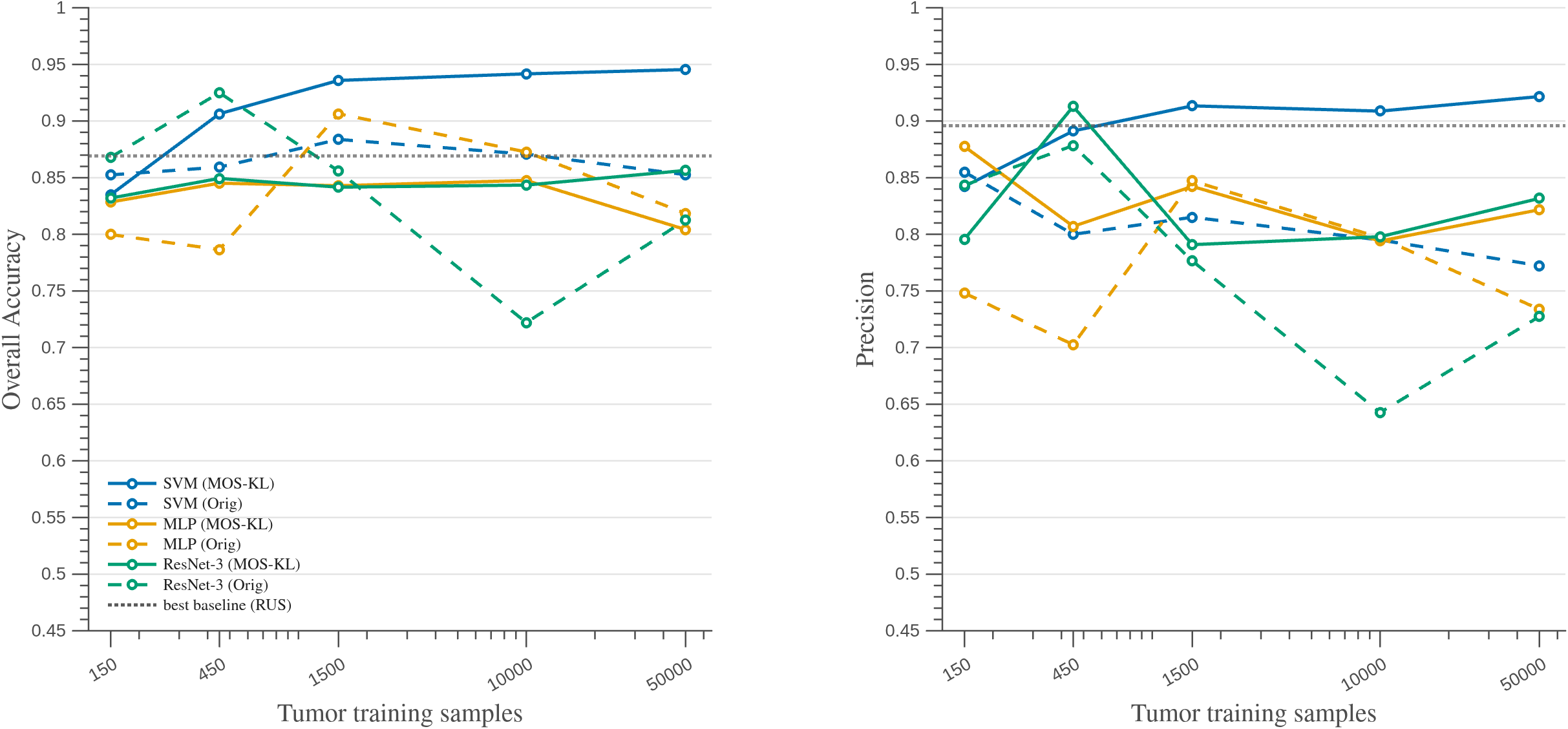}
  \caption{$\sin(18 x)$ --- Accuracy and precision comparison results as
    the number of samples in the training datasets increases for class
    $\bA$ $(\mcA_{{N_\bA}})$.}
  \label{fig:semisyn18}
\end{figure} 

\textbf{Test \#2:} We increase the difficulty by deforming the original
realizations with a higher-frequency sine, $u^{\bA}_M(\bx,\omega_k) \leftarrow
\sin(18\, u^{\bA}_M(\bx,\omega_k))$ and $u^{\bB}_M(\bx,\omega_k) \leftarrow
\sin(18\, u^{\bB}_M(\bx,\omega_k))$, and repeat the sweep over $N_{\bA}$
(Table~\ref{tab:semisyn_sin18}, Figure~\ref{fig:semisyn18}). The qualitative
behavior of Test~\#1 is preserved, but the separation between the multilevel and
original features widens. The MOS-KL + SVM RBF accuracy again climbs with the data, from $83.5\%$ to $94.5\%$, while the original-feature SVM stays below 88.4\% and its specificity collapses. At $N_{\bA}=50{,}000$
the multilevel SVM reaches $94.5\%$ accuracy versus $85.3\%$ on the original features, and exceeds the best ensemble baseline ($91.2\%$).

\begin{table*}[t]
\centering
\small
\setlength{\tabcolsep}{4.5pt}
\renewcommand{\arraystretch}{1.15}
\begin{tabular}{@{}l l >{\columncolor{mlsbg}}c c >{\columncolor{mlsbg}}c c >{\columncolor{mlsbg}}c c >{\columncolor{mlsbg}}c c c@{}}
\toprule
& & \multicolumn{2}{c}{SVM} & \multicolumn{2}{c}{MLP} & \multicolumn{2}{c}{ResNet-3} & \multicolumn{2}{c}{ResNet-30} & \\
\cmidrule(lr){3-4}\cmidrule(lr){5-6}\cmidrule(lr){7-8}\cmidrule(lr){9-10}
\multicolumn{1}{c}{Tumor} & Metric & MOS-KL & Orig & MOS-KL & Orig & MOS-KL & Orig & MOS-KL & Orig & \multirow{-2}{*}{Baseline\textsuperscript{a}} \\
\midrule
\multirow{4}{*}{150} & Accuracy & 0.778\textsuperscript{L7} & 0.795 & 0.787\textsuperscript{L0} & 0.777 & 0.798\textsuperscript{L1} & \textbf{0.869} & 0.692\textsuperscript{L0} & 0.564 & 0.792\,\textsuperscript{RUS} \\
 & Sensitivity & 0.759 & 0.807 & 0.770 & 0.871 & 0.846 & \textbf{0.927} & 0.643 & 0.846 & 0.801 \\
 & Specificity & 0.798 & 0.783 & 0.803 & 0.684 & 0.750 & \textbf{0.810} & 0.741 & 0.283 & 0.783 \\
 & AUC & 0.849 & 0.874 & 0.856 & 0.885 & 0.888 & \textbf{0.944} & 0.757 & 0.659 & 0.747 \\
\midrule
\multirow{4}{*}{450} & Accuracy & \textbf{0.873}\textsuperscript{L7} & 0.800 & 0.790\textsuperscript{L5} & 0.729 & 0.791\textsuperscript{L6} & 0.759 & 0.693\textsuperscript{L4} & 0.637 & 0.843\,\textsuperscript{RUS} \\
 & Sensitivity & 0.865 & 0.950 & 0.844 & 0.986 & 0.755 & \textbf{0.995} & 0.676 & 0.930 & 0.904 \\
 & Specificity & \textbf{0.881} & 0.649 & 0.737 & 0.471 & 0.827 & 0.522 & 0.710 & 0.344 & 0.782 \\
 & AUC & 0.944 & 0.937 & 0.871 & 0.950 & 0.870 & \textbf{0.954} & 0.753 & 0.815 & 0.923 \\
\midrule
\multirow{4}{*}{1,500} & Accuracy & \textbf{0.908}\textsuperscript{L7} & 0.814 & 0.761\textsuperscript{L5} & 0.680 & 0.779\textsuperscript{L7} & 0.695 & 0.656\textsuperscript{L6} & 0.586 & 0.860\,\textsuperscript{RUS} \\
 & Sensitivity & 0.909 & 0.990 & 0.698 & 0.997 & 0.801 & \textbf{1.000} & 0.677 & 0.999 & 0.911 \\
 & Specificity & \textbf{0.908} & 0.638 & 0.824 & 0.363 & 0.758 & 0.389 & 0.634 & 0.174 & 0.809 \\
 & AUC & 0.968 & 0.955 & 0.852 & 0.977 & 0.845 & \textbf{0.995} & 0.705 & 0.915 & 0.940 \\
\midrule
\multirow{4}{*}{10,000} & Accuracy & \textbf{0.919}\textsuperscript{L7} & 0.789 & 0.769\textsuperscript{L4} & 0.710 & 0.781\textsuperscript{L6} & 0.621 & 0.630\textsuperscript{L2} & 0.594 & 0.852\,\textsuperscript{RUS} \\
 & Sensitivity & 0.937 & 0.999 & 0.880 & 0.999 & 0.899 & \textbf{1.000} & 0.654 & 0.999 & 0.909 \\
 & Specificity & \textbf{0.901} & 0.579 & 0.657 & 0.421 & 0.663 & 0.242 & 0.605 & 0.189 & 0.796 \\
 & AUC & 0.974 & 0.944 & 0.841 & 0.948 & 0.860 & \textbf{0.999} & 0.674 & 0.953 & 0.936 \\
\midrule
\multirow{4}{*}{50,000} & Accuracy & \textbf{0.920}\textsuperscript{L7} & 0.740 & 0.741\textsuperscript{L6} & 0.530 & 0.751\textsuperscript{L5} & 0.533 & 0.606\textsuperscript{L3} & 0.727 & 0.857\,\textsuperscript{RUS} \\
 & Sensitivity & 0.925 & \textbf{1.000} & 0.856 & \textbf{1.000} & 0.735 & \textbf{1.000} & 0.691 & \textbf{1.000} & 0.930 \\
 & Specificity & \textbf{0.915} & 0.479 & 0.626 & 0.060 & 0.767 & 0.065 & 0.522 & 0.455 & 0.784 \\
 & AUC & 0.974 & 0.935 & 0.816 & 0.664 & 0.828 & 0.879 & 0.645 & \textbf{0.997} & 0.942 \\
\bottomrule
\end{tabular}
\caption{GCM semisynthetic, $\sin(20 x)$ --- same setup as Table~\ref{tab:semisyn_sin15}.}
\label{tab:semisyn_sin20}
\end{table*} 

\begin{figure}[h]
  \centering
  \includegraphics[scale = 0.5]{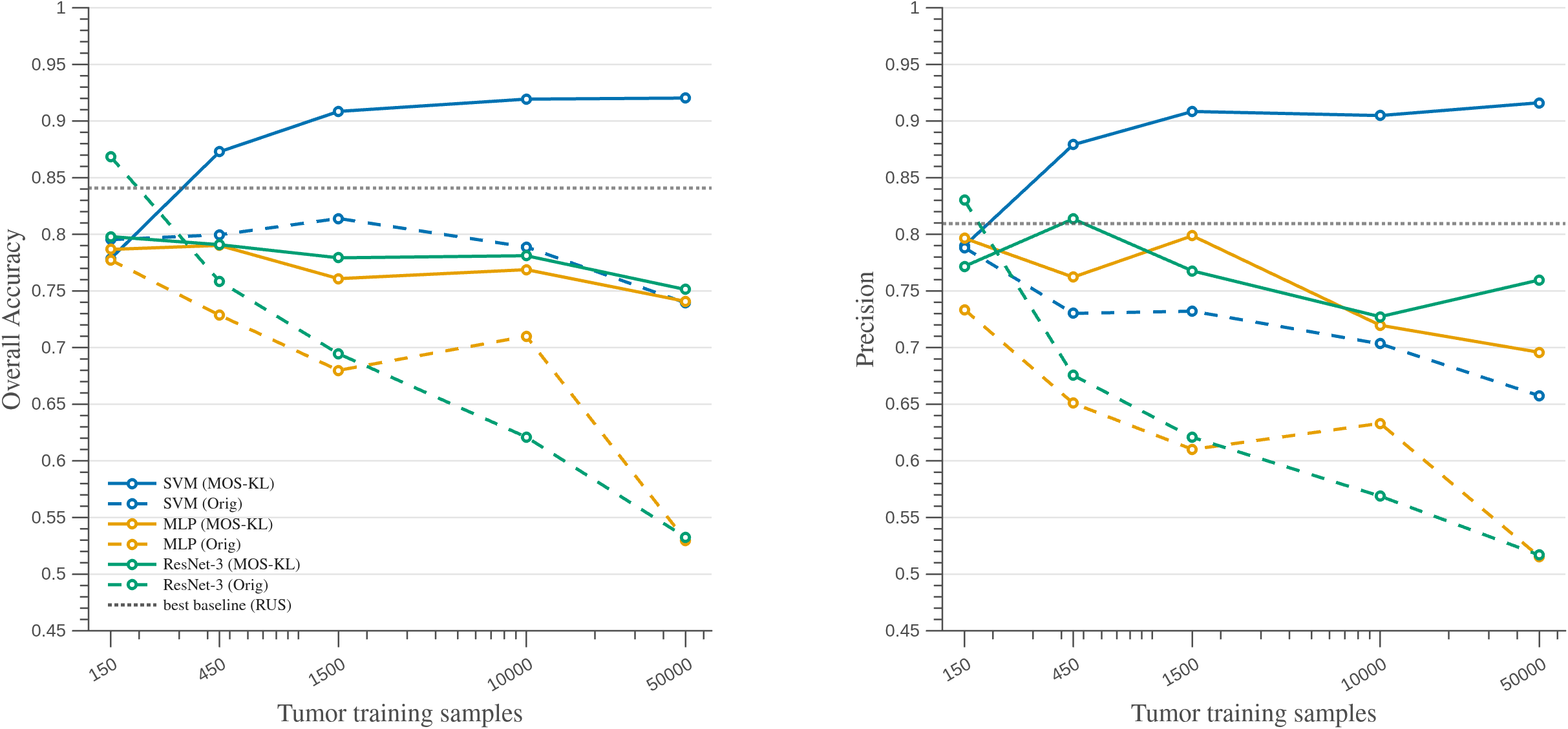}
  \caption{$\sin(20 x)$ --- Accuracy and precision comparison results as
    the number of samples in the training datasets increases for class
    $\bA$ $(\mcA_{{N_\bA}})$.}
  \label{fig:semisyn20}
\end{figure} 

\textbf{Test \#3:} Finally we apply the strongest deformation,
$u^{\bA}_M(\bx,\omega_k) \leftarrow \sin(20\, u^{\bA}_M(\bx,\omega_k))$ and
$u^{\bB}_M(\bx,\omega_k) \leftarrow \sin(20\, u^{\bB}_M(\bx,\omega_k))$
(Table~\ref{tab:semisyn_sin20}, Figure~\ref{fig:semisyn20}). This is the hardest
task, and it is where the multilevel advantage is largest. The MOS-KL + SVM RBF
accuracy rises from $77.8\%$ to $92.0\%$ with $N_{\bA}$, whereas the
original-feature SVM decays sharply from $79.5\%$ to $74.0\%$; and it also surpasses the best ensemble baseline ($85.7\%$). At $N_{\bA}=50{,}000$ MOS-KL + MLP yields a $74.1\%$ accuracy while MLP alone degenerates to random guess. 
At the smallest size the deep original-feature ResNet-3 briefly leads ($86.9\%$ against the multilevel SVM's $77.8\%$), but it cannot exploit the additional data, and the multilevel SVM overtakes it already at $N_{\bA}=450$ and dominates thereafter.

An important observation is that the MOS-KL features are substantially more robust
to the deformation than any method on the original features. As the frequency of
the sine transform is increased from $15$ to $20$, the maximum MOS-KL SVM accuracy
decreases only slowly, from $97.3\%$ to $94.5\%$ to $92.0\%$, whereas the
original-feature SVM decays much faster, from $92.2\%$ to $85.3\%$ to $74.0\%$; the
tree ensembles behave similarly. The gap between the multilevel and original
representations therefore grows both with the training-set size and with the
difficulty of the task.

\begin{remark}
  One key conclusion from these tests is that as the number of realizations is
  increased, the accuracy of the MOS-KL SVM RBF classifier generally increases.
  This is expected, since the estimate of the covariance function becomes better
  and the separation of the classes improves (Theorem~\ref{mls:theo3}). This is a
  good approach to the difficult problem of imbalanced datasets: performance
  improves rather than degrades as the imbalance grows, because the surplus
  majority-class data are diverted to the covariance construction rather than
  discarded. This is in contrast to methods that balance the data by subsampling,
  which lose information, or by bootstrap resampling, which can be unreliable. The
  tree ensembles (Gradient Boosting and Random Forest) are the most sensitive to
  the imbalance and collapse toward the majority class at the largest sizes, while
  SVM RBF with the posterior option and RUS Boost remain comparatively robust.
\end{remark}

\bibliographystyle{plain}
\bibliography{citations,multilevel,changedetectionreferences,FDA,KonRefs}

\end{document}